\documentclass[11pt]{article}

\PassOptionsToPackage{numbers, compress}{natbib}
\usepackage{natbib}

\usepackage{hyperref}       %

\hypersetup{
    colorlinks=true,
    linkcolor=blue,
    urlcolor=magenta,
    citecolor=blue}

\date{} %
\advance \oddsidemargin by -0.8in
\newcommand{\myparskip}{3pt}
\parskip \myparskip

\usepackage[utf8]{inputenc} %
\usepackage[T1]{fontenc}    %
\usepackage{algorithm}
\usepackage{algpseudocode}
\usepackage{xspace}
\usepackage{multirow}
\usepackage{pifont}
\usepackage{xcolor}
\usepackage{booktabs}
\usepackage{color} 
\usepackage{diagbox}
\usepackage{graphicx}
\usepackage{csquotes}
\usepackage{siunitx}
\usepackage{stfloats}
\usepackage{amsmath}
\usepackage{amsfonts}
\usepackage{bbm}
\usepackage{makecell}
\usepackage[usestackEOL]{stackengine}

\usepackage{lineno}

\newcommand{\x}{\mathbf x}
\newcommand{\y}{\mathbf y}
\newcommand{\f}{f}
\newcommand{\imp}{\mathbf \phi}
\newcommand{\phe}{\Phi}

\newcommand{\eps}{\mathbf \epsilon}
\DeclareMathOperator*{\argmax}{arg\,max}
\newcommand{\topkimpindices}{\mathbbm{1}}

\newif\ifcomments
\ifcomments
    \providecommand{\sameer}[2][]{{\protect\color{magenta}{[sameer:\textbf{#1} #2]}}}
    \providecommand{\hima}[2][]{{\protect\color{red}{[Hima:\textbf{#1} #2]}}}
    \providecommand{\satya}[2][]{{\protect\color{magenta}{[satya:\textbf{#1} #2]}}}
    \providecommand{\dylan}[2][]{{\protect\color{blue}{[dylan:\textbf{#1} #2]}}}
\else
    \providecommand{\hima}[2][]{}
    \providecommand{\sameer}[2][]{}
    \providecommand{\dylan}[2][]{}
    \providecommand{\satya}[2][]{}
\fi

\newcommand{\sys}{TalkToModel\xspace}

\definecolor{textexample}{rgb}{0.23, 0.30, 0.45}
\newcommand{\userinput}[1]{\textcolor{textexample}{``#1''\xspace}}
\newcommand{\openui}[1]{\textcolor{textexample}{``#1\xspace}}
\newcommand{\closeui}[1]{\textcolor{textexample}{#1''\xspace}}
\newcommand{\midui}[1]{\textcolor{textexample}{#1\xspace}}

\definecolor{parsecolor}{rgb}{0.0, 0.0, 0.4}
\newcommand{\parse}[1]{\textcolor{parsecolor}{\texttt{#1}\xspace}}

\definecolor{argcolor}{rgb}{0.54, 0.0, 0.0}
\newcommand{\args}[1]{\textcolor{argcolor}{\texttt{#1}\xspace}}

\definecolor{lightgrey}{rgb}{0.859, 0.859, 0.859}
\newcommand{\sysresponse}[1]{\colorbox{lightgrey} {#1}}

\definecolor{lightblue}{rgb}{0.09, 0.51, 0.98}

\newcommand{\myquad}[1][1]{\hspace*{#1em}\ignorespaces}

\newcommand{\wildcard}[1]{\texttt{\textcolor{blue}{\{{#1}\}}}\xspace}

\makeatletter
\newcommand{\printfnsymbol}[1]{%
  \textsuperscript{\@fnsymbol{#1}}%
}
\makeatother

\AtBeginDocument{%
  \providecommand\BibTeX{{%
    \normalfont B\kern-0.5em{\scshape i\kern-0.25em b}\kern-0.8em\TeX}}}

\usepackage{graphicx}
\title{Explaining Machine Learning Models with Interactive Natural Language Conversations Using TalkToModel}
\author{%
  Dylan Slack \\
  UC Irvine \\
  \texttt{dslack@uci.edu} \\
  \and
  Satyapriya Krishna \\
  Harvard University \\
  \texttt{skrishna@g.harvard.edu} \\
  \and
  Himabindu Lakkaraju\thanks{Equal Contribution} \\
  Harvard University \\
  \texttt{hlakkaraju@hbs.edu} \\
  \and
  Sameer Singh\printfnsymbol{1} \\
  UC Irvine / AI2 \\
  \texttt{sameer@uci.edu} \\
 }

\begin{document}

\maketitle

\begin{abstract}
\dylan{up to 150 words}
Practitioners increasingly use machine learning (ML) models, yet they have become more complex and harder to understand.
To address this issue, researchers have proposed techniques to explain model predictions.
However, practitioners struggle to use explainability methods because they do not know which to choose and how to interpret the results.
We address these challenges by introducing \sys: an interactive dialogue system that enables users to explain ML models through natural language conversations. \sys comprises three components: \textbf{1)} an adaptive dialogue engine that interprets natural language and generates meaningful responses,
\textbf{2)} an execution component, which constructs the explanations used in the conversation, \textbf{3)} a conversational interface.
In real-world evaluations, $73\%$ of healthcare workers agreed they would use \sys over existing systems for understanding a disease prediction model, and $85\%$ of ML professionals agreed \sys was easier to use, demonstrating that \sys is highly effective for model explainability.
\end{abstract}

\section{Introduction}
Due to their strong performance, machine learning (ML) models increasing make consequential decisions in several critical domains, such as healthcare, finance, and law.
However, state-of-the-art ML models, such as deep neural networks, have become more complex and hard to understand.
This dynamic poses challenges in real-world applications for model stakeholders who need to understand why models make predictions and whether to trust them.
Consequently, practitioners have often turned to \textit{inherently interpretable} machine learning models for these applications including decision lists and sets~\cite{lakkaraju2016interpretable, angelino2017learning, ustunSuperSparse2016, zengRecid2015} and generalized additive models~\cite{lou2013accurate, NEURIPS2021_251bd044, chang2021node, chang2021interpretable, zhangaxiomatic}, which people can more easily understand.
Nevertheless, black-box models are often more flexible and accurate, motivating the development of \textit{post-hoc} explanations that explain the predictions of trained ML models.
These techniques either fit faithful models in the local region around a prediction or inspect internal model details, like gradients, to explain predictions~\cite{lime:whi16, slackreliable21, gradcam, slack2021defuse, Ribeiro_Singh_Guestrin_2018, oodhase2021, Simonyan14a, smilkov2017smoothgrad}.

Yet, recent work suggests practitioners often have difficulty using explainability techniques~\cite{lakkaraju2022rethinking, interinterpKaur2020, weldintelligible}.
These challenges are due to difficulty figuring out which explanations to implement, how to interpret their results, and answering followup questions beyond the initial explanation.
In the past, researchers have proposed several point-and-click dashboard techniques to help overcome these issues, such as the Language Interpretability Tool (LiT)~\cite{tenney2020language}, which is designed to understand natural language processing (NLP) models and the "What-If" Tool~\cite{whatif}—a tool aimed at performing counterfactual analyses for models. 
However, these methods still require a high level of expertise because users must know which explanations to run and lack the flexibility to support arbitrary follow up questions users might have.
Overall, being able to understand ML models through simple and intuitive interactions is a key bottleneck in adoption across many applications.

Natural language dialogues are a promising solution for supporting broad and accessible interactions with ML models due to their ease of use, capacity, and support for continuous discussion.
However, designing a dialogue system that enables a satisfying model understanding experience introduces several challenges.
First, the system must handle many conversation topics about the model and data while facilitating natural conversation flow~\cite{Ward2015TenCI}.
For instance, these topics may include explainability questions like the most important features for predictions and general questions such as data statistics or model errors.
Further, the system must work for a variety of model classes and data, and it should understand language usage across different settings~\cite{Carenini1994-wc}.
For example, participants will use different terminology in conversations about loan prediction compared to disease diagnosis.
Last, the dialogue system should generate accurate responses that address the users' core questions~\cite{Pennebaker2002-sq, Zhang2020-lv}.
In the literature, researchers have suggested some prototype designs for generating explanations using natural language.
However, these initial designs address specific explanations and model classes, limiting their applicability in general conversational explainability settings~\cite{sokol2018glass, feldhus2022mediators}.

\begin{figure}[t]
    \centering
    \includegraphics[trim={5.5cm 0.2cm 5.5cm 0.65cm},clip,width=.7\columnwidth]{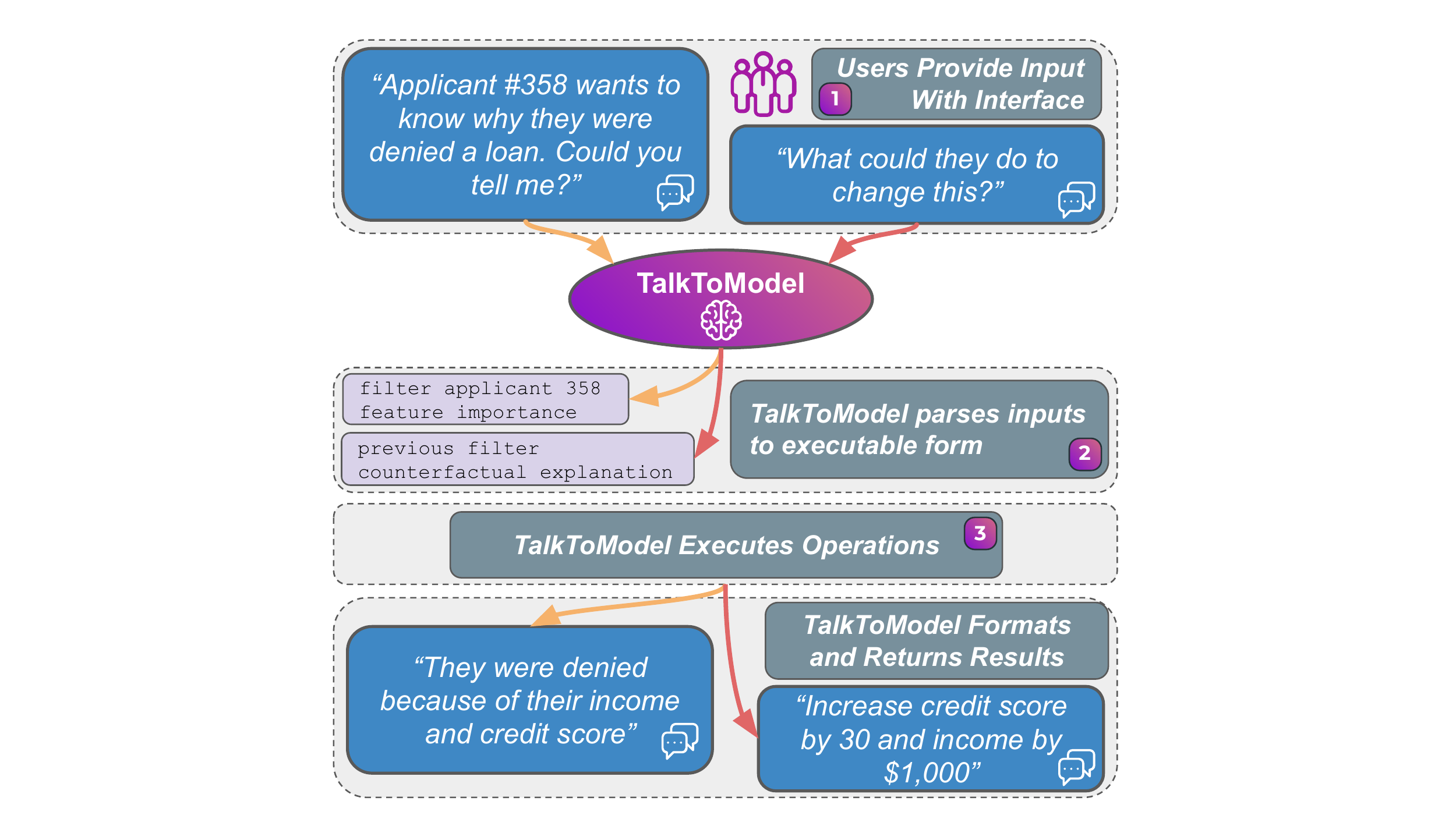}
    \caption{\textbf{Overview of \sys:} Instead of writing code, users have conversations with \sys as follows:
    \textbf{(1)} users supply natural language inputs.
    \textbf{(2)} the dialogue engine parses the input into an executable representation.
    \textbf{(3)} the execution engine runs the operations and the dialogue engine uses the results in its response.
    }
    \label{fig:fig1}
\end{figure}

In this work, we address these challenges by introducing \sys: a system that enables open-ended natural language dialogues for understanding ML models for any tabular dataset and classifier
(an overview of \sys is provided in Figure~\ref{fig:fig1}).
Users can have discussions with \sys about why predictions occur, how the predictions would change if the data changes, and how to flip predictions, among many other conversation topics (an example conversation is provided in Table~\ref{tab:conversation}).
Further, they can perform these analyses on any group in the data, such as a single instance or a specific group of instances.
For example, on a disease prediction task, users can ask \userinput{how important is BMI for the predictions?} or \userinput{so how would decreasing the glucose levels by ten change the likelihood of men older than twenty having the disease?}
\sys will respond by describing how, for instance, BMI is the most important feature for predictions, and decreasing glucose will decrease the chance by $20\%$.
From there, users can engage further in the conversation by asking follow up questions like, \userinput{what if you instead increased glucose by ten for that group of men?} and \sys use the context to accurately respond.
Conversations with \sys make model explainability straightforward because users can simply talk with the system in natural language about the model, and the system will generate useful responses.

To support such rich conversations with \sys, we introduce techniques for both language understanding and model explainability. 
First, we propose a \textit{dialogue engine} that parses user text inputs (referred to as \textit{user utterances}) into an SQL-like programming language using a large language model (LLM).
The LLM performs the parsing by treating the task of translating user utterances into the programming language as a seq2seq learning problem, where the user utterances are the source and parses in the programming language are the targets~\cite{sutskever2014}.
In addition, the \sys language combines operations for explanations, ML error analyses, data manipulation, and descriptive text into a single language capable of representing a wide-variety of potential conversation topics most model explainability needs (an overview of the different operations is provided in Table~\ref{tab:operations}).
To support the system adapting to any dataset and model, we introduce lightweight adaption techniques to fine-tune LLMs to perform the parsing, enabling strong generalization to new settings.
 Second, we introduce an \textit{execution engine} that runs the operations in each parse.
To reduce the burden of users deciding which explanations to run, we introduce methods that automatically select explanations for the user.
In particular, this engine runs many explanations, compares their fidelities, and selects the most accurate ones.
Finally, we construct a \textit{text interface} where users can engage in open-ended dialogues using the system, enabling anyone, including those with minimal technical skills, to understand ML models.

\newcommand{\placeresponse}[1]{\sysresponse{#1}}
\newcommand{\rightalignthis}[1]{\myquad[3]{#1}}
\newcommand{\uiONE}{\Centerstack[l]{\openui{What are the predictions on patients older than 30? Could you}\\\midui{also tell me what would happen to these predictions if we}\\\closeui{increased bmi by 10?}}}
\newcommand{\rONE}{\Centerstack[l]{ On data with \textbf{age greater than 30.0}, the model predicts: \\  \quad\ding{233} \textbf{No Diabetes} on 46.8\% of data \\  \quad\ding{233} \textbf{Diabetes} on 53.2\% of data \\ On data with \textbf{age greater than 30.0 and bmi is increased \textbf{by 10.0,}}\\ the model predicts:\\ \quad\ding{233} \textbf{No Diabetes} on 30.1\% of data \\ \quad\ding{233} \textbf{Diabetes} on 69.9\% of data}}
\newcommand{\uiTWO}{\Centerstack[l]{\openui{So how important is bmi for the model's predictions on the}\\ \closeui{over thirty group?}}}
\newcommand{\rTWO}{\Centerstack[l]{For the model's predictions on data with \textbf{age greater than 30.0}, the \\ \textbf{bmi} feature  is ranked on average \textbf{2.86} (95\% CI [2.685, 3.035]). Here,\\rank 1 is the most important feature (out of 8 features). Overall, \textbf{bmi}\\is \textbf{highly important} for the model's predictions.}}
\newcommand{\uiTHREE}{\Centerstack[l]{\openui{I want to learn more about patient 378, please tell me the model's}\\\closeui{prediction and why it occurred.} }}
\newcommand{\rTHREE}{\Centerstack[l]{The instance with \textbf{id equal to 378} is predicted to \textbf{have diabetes.} For\\this prediction:\\\quad\ding{233} \textbf{glucose} is the most important feature and has a \textit{positive} influence\\\quad\ding{233} \textbf{bmi} is the second most important feature and has a \textit{positive} influence\\\quad\ding{233} \textbf{diabetes pedigree function} is the third most important feature and\\\myquad[2] has a \textit{positive} influence\\This explanation fits the model with R2 of 0.81, meaning it is \textbf{accurate}.\\ Want to see the feature importance scores?}}
\newcommand{\uiFOUR}{\Centerstack[l]{\openui{On this data, what's the model's sensitivity and specificity?}}}
\newcommand{\rFOUR}{\Centerstack[l]{The model scores \textit{0.667 sensitivity} on the data where \textbf{age greater}\\ \textbf{than 30.0.} \\ The model scores \textit{0.638 specificity} on the data where \textbf{age greater} \\ \textbf{than 30.0.}}}
\newcommand{\uiFIVE}{\Centerstack[l]{\openui{What types of patients is the model typically predicting}\\\closeui{incorrect here}?}}
\newcommand{\rFIVE}{\Centerstack[l]{For data with \textbf{age greater than 30.0}, the model typically predict\\predicts incorrect:\\\quad\ding{233} if (\textbf{bmi > 26.95}) and \textbf{(glucose $\leq$ 125.0)} and \textbf{(pedigree\_function }\\\myquad[2]then the model is incorrect on \textit{62.3\%} of \textit{53} samples\\\quad\ding{233} \textbf{(bmi > 26.95)} and \textbf{(glucose > 125.0)} and \textbf{(age > 57.5)}\\\myquad[2]\textbf{<= 0.943}) then the model is incorrect on \textit{62.5\%} of \textit{8} samples\\\quad\ding{233} \textbf{(bmi $\leq$ 26.95)} and \textbf{(glucose > 180.5)} then the model is incorrect\\\myquad[2]on \textit{100.0\%} of \textit{2} samples.\\Want to take a closer look at these rules?}}
\newcommand{\uiSIX}{\Centerstack[l]{\openui{Ah, so which two features are mostly more important than}\\\closeui{bmi for these people?}}}
\newcommand{\rSIX}{\Centerstack[l]{For the models predictions on data \textbf{where age is greater than 30.0},\\the \textbf{top 2} most important features are:\\\quad1. \textbf{glucose}\\\quad2. \textbf{pedigree\_function}}}
\newcommand{\uiSEVEN}{\Centerstack[l]{\userinput{Not right now. Could you instead show me how to flip this prediction?}}}
\newcommand{\rSEVEN}{\Centerstack[l]{For the instance with \textbf{id equal to 378},\\ \quad\ding{233} If you \textit{decrease glucose to 85.0} and \textit{decrease bmi to 31.2}, the model\\\myquad[2]will predict \textbf{no diabetes.}\\\quad\ding{233} Also, if you \textit{decrease glucose to 12.0} and \textit{decrease bloodpressure to},\\ \myquad[2]31.0 the model will predict \textbf{no diabetes}.\\Want to see a couple more options?}}
\newcommand{\uiEIGHT}{\Centerstack[l]{\userinput{Actually, could you remind me of this patients glucose and bmi levels?}}}
\newcommand{\rEIGHT}{\Centerstack[l]{For the instance with \textbf{id equal to 378}, the value of \textbf{glucose} is \textit{156}.\\For the instance with \textbf{id equal to 378}, the value of \textbf{bmi} is \textit{48.3}.}}
\newcommand{\uiNINE}{\Centerstack[l]{\userinput{Was the model's prediction correct for this patient?}}}
\newcommand{\rNINE}{\Centerstack[l]{For the instance with \textbf{id equal to 378}, the model predicts correctly.}}
\newcommand{\uiTEN}{\Centerstack[l]{\userinput{What's the distribution of glucose ?}}}
\newcommand{\rTEN}{\Centerstack[l]{For the instance with \textbf{id equal to 378}, the model predicts correctly.}}
\newcommand{\uieleven}{\Centerstack[l]{\openui{Could you show me the likelihood if this person having diabetes if}\\\closeui{you changed their glucose to eighty-five?}}}
\newcommand{\releven}{\Centerstack[l]{The model predicts the instance with \textbf{id equal to 378} and \textbf{glucose is set}\\\textbf{to 85.0} as:\\\quad\ding{233} \textbf{No Diabetes} with \textit{92.3}\% probability\\\quad\ding{233} \textbf{Diabetes} with \textit{7.7}\% probability}}
\newcommand{\uiTWELVE}{\Centerstack[l]{\userinput{Did the model get this prediction correct?}}}
\newcommand{\rTWELVE}{\Centerstack[l]{For the data with \textbf{id equal to 378}, the model predicts correctly!}}
\newcommand{\uiTHIR}{\Centerstack[l]{\openui{What's the model's positive predictive value on people with}\\\closeui{bmi's greater than 45 and glucose over 150?}}}
\newcommand{\rTHIR}{\Centerstack[l]{The model scores 0.571 ppv on the data where \textbf{bmi greater than 45.0}\\\textbf{and glucose greater than 150.0.}}}
\newcommand{\uiFOURTEEN}{\Centerstack[l]{\userinput{Show me examples of a few mistakes from that group?}}}
\newcommand{\rFOURTEEN}{\Centerstack[l]{For the data with \textbf{bmi greater than 45.0 and glucose greater than 150.0},\\the model is \textit{incorrect 3 out of 7 times (error rate 0.429)}. Here are the\\ids of instances the model predicts incorrectly: [487 469 247]. Do you\\want to see any?}}

\begin{table*}[h!]
    \small
    \centering
    \caption{A conversation about diabetes prediction, demonstrating the breadth of different conversation points the system can discuss.\vspace{1mm}}
    \label{tab:conversation}
    \begin{tabular}{ll}
        \toprule
        \multicolumn{1}{c}{User \userinput{utterances} \& \sys \sysresponse{response}}
         \\
         \midrule
         \vspace{1mm}\rightalignthis{\uiONE} \\
         \vspace{1.0mm}\placeresponse{\rONE}  \\
         \vspace{1mm}\rightalignthis{\uiTWO}  \\
         \vspace{0.5mm}\placeresponse{\rTWO} \\
         \vspace{0.5mm}\rightalignthis{\uiSIX} \\
         \vspace{1mm}\placeresponse{\rSIX} \\
         \vspace{0.5mm}\rightalignthis{\uiFIVE} \\
         \vspace{1mm}\placeresponse{\rFIVE} \\
        \midrule
    \end{tabular}
\end{table*}
\begin{table}[h!]
    \centering
    \caption{Overview of the \textit{operations} supported by \sys, which are incorporated into the conversation to generate responses.\vspace{1mm}}
    \label{tab:operations}
    \begin{tabular}{ll}
     \toprule
        & \bf \parse{operation}, \args{arguments}, and description \\ \midrule
        \vspace{.5mm} \multirow{10}{*}{\centering \rotatebox[origin=c]{90}{Data}} & \Centerstack[l]{\parse{filter(}\args{dataset, feature, value, comparison}\parse{)}: filters\\ \args{dataset} by using value and comparison operator} \\
        \vspace{.5mm} &  \Centerstack[l]{\parse{change(}\args{dataset, feature,} \args{ value, variation}\parse{)}: Changes\\\args{dataset} by increasing, decreasing, or setting feature by \args{value}} \\
        \vspace{.5mm} & \parse{show(}\args{list}\parse{)}: Shows items in list in the conversation \\
        \vspace{.5mm} & \Centerstack[l]{ \parse{statistic(}\args{dataset, metric, feature}\parse{)}: Computes\\summary statistic for \args{feature}} \\
        \vspace{.5mm} & \parse{count(}\args{list}\parse{)}: Length of list \\
        \vspace{.5mm} & \parse{and(}\args{op1, op2}\parse{)}: Logical ``and'' of two operations  \\
        \vspace{.5mm} & \parse{or(}\args{op1, op2}\parse{)}: Logical ``or'' of two operations \\ \midrule
        \vspace{.5mm} \multirow{7}{*}{\centering \rotatebox[origin=c]{90}{Explainability}} & \parse{explain(}\args{dataset, method, class=predicted}\parse{)}: Feature importances on \args{dataset} \\
        \vspace{.5mm} & \parse{cfe(}\args{dataset, number, class=opposite}\parse{)}: Gets \args{number} counterfactual explanations \\
        \vspace{.5mm} & \parse{topk(}\args{dataset, k}\parse{)}: Top \args{k} most important features \\
        \vspace{.5mm} & \parse{important(}\args{dataset, feature}\parse{)}: Importance ranking of \args{feature} \\
        \vspace{.5mm} & \parse{interaction(}\args{dataset}\parse{)}: Interaction effects between features \\
        \vspace{.5mm} & \parse{mistakes(}\args{dataset}\parse{)}: Patterns in the model's errors on \args{dataset}   \\ \midrule
        \vspace{.5mm} \multirow{4}{*}{\centering \rotatebox[origin=c]{90}{ML}} & \parse{predict(}\args{dataset}\parse{)}: Model predictions on \args{dataset} \\
        \vspace{.5mm} & \parse{likelihood(}\args{dataset}\parse{)}: Prediction probabilities on \args{dataset} \\
        \vspace{.5mm} & \parse{incorrect(}\args{dataset}\parse{)}: Incorrect predictions \\
        \vspace{.5mm} & \parse{score(}\args{dataset, metric}\parse{)}: Scores the model with \args{metric} \\ \midrule
        \vspace{.5mm} \multirow{3}{*}{\centering \rotatebox[origin=c]{90}{Conv.}} & \parse{prev\_filter(}\args{conversation}\parse{)}: Gets last filters \\
        \vspace{.5mm} & \parse{prev\_operation(}\args{conversation}\parse{)}: Gets last non-filtering operations \\
        
        \vspace{.5mm} & \parse{followup(}\args{conversation}\parse{)}: Respond to system followups \\\midrule
        \vspace{.5mm} \multirow{4}{*}{\centering \rotatebox[origin=c]{90}{Description}} & \parse{function(}\args{}\parse{)}: Overview of the system's capabilities \\
        \vspace{.5mm} & \parse{data(}\args{dataset}\parse{)}: Summary of dataset \\
        \vspace{.5mm} & \parse{model(}\args{}\parse{)}: Description of \args{model} \\ 
        \vspace{.5mm} & \parse{define(}\args{term}\parse{)}: Defines \args{term} \\ \bottomrule
    \end{tabular}
\end{table}

\section{Results} 
\label{sec:eval}

\begin{table*}[ht!]
    \centering
        \caption{
        {Exact Match Parsing Accuracy (\%) for the $3$ gold datasets, on the IID and Compositional splits, as well as Overall. The fine-tuned T5 models perform significantly better than few-shot GPT-J, and T5 Large performed the best. These results demonstrate that \sys can understand user intentions with a high degree of accuracy using the T5 models.\vspace{1mm}}
    }
    \label{tab:parsing_results}
    \begin{tabular}{lccccccccc}
        \toprule
        & \multicolumn{3}{c}{German} & \multicolumn{3}{c}{Compas} & \multicolumn{3}{c}{Diabetes}
        \\
        \cmidrule(lr){2-4} \cmidrule(lr){5-7} \cmidrule(lr){8-10}
        & IID & Comp. & Overall & IID & Comp. & Overall & IID & Comp. & Overall \\
        \midrule
        Nearest Neighbors      & 26.2 & 0.0 & 16.5  & 27.4 & 0.0 & 21.9  & 10.9 & 0.0 & 8.4  \\
        \midrule
        GPT-Neo 1.3B           &  &  &  &  &  &  &  &  &  \\
                \hspace{0.2em} \textsc{10-shot}  & 41.3 & 4.1 & 27.5  & 35.9 & 0.0 & 28.8  & 40.1 & 7.0 & 32.6  \\
                \hspace{0.2em} \textsc{20-shot}  & 39.7 & 0.0 & 25.0  & 39.3 & 0.0 & 31.5  & 42.9 & 2.3 & 33.7  \\
                \hspace{0.2em} \textsc{30-shot}  & 42.9 & 0.0 & 27.0  & 39.3 & 0.0 & 31.5  & 41.5 & 4.7 & 33.2  \\
        \midrule
        GPT-Neo 2.7B           &  &  &  &  &  &  &  &  &  \\
                \hspace{0.2em} \textsc{5-shot}  & 38.1 & 4.1 & 25.5  & 35.9 & 3.4 & 29.5  & 46.9 & 7.0 & 37.9  \\
                \hspace{0.2em} \textsc{10-shot}  & 38.1 & 6.8 & 26.5  & 40.2 & 3.4 & 32.9  & 40.8 & 9.3 & 33.7  \\
                \hspace{0.2em} \textsc{20-shot}  & 39.7 & 0.0 & 25.0  & 39.3 & 0.0 & 31.5  & 42.9 & 2.3 & 33.7  \\
        \midrule
        \textsc{GPT-J 6B}           &  &  &  &  &  &  &  &  &  \\
                \hspace{0.2em} \textsc{5-shot}  & 51.6 & 14.9 & 38.0  & 51.3 & 6.9 & 42.5  & 55.8 & 7.0 & 44.7  \\
                \hspace{0.2em} \textsc{10-shot}  & 57.9 & 9.5 & 40.0  & 49.6 & 3.4 & 40.4  & 53.7 & 9.3 & 43.7  \\
        \midrule
        \textsc{T5}           &  &  &  &  &  &  &  &  &  \\
                \hspace{0.2em} \textsc{small}  & 61.1 & 32.4 & 50.5  & 71.8 & 10.3 & 59.6  & 77.6 & 30.2 & 66.8  \\
                \hspace{0.2em} \textsc{base}  & 68.3 & \bf 48.6 & 61.0  & 65.0 & 10.3 & 54.1  & \bf 84.4 & 34.9 & 73.2  \\
                \hspace{0.2em} \textsc{large}  & \bf 74.6 & 44.6 & \bf 63.5  & \bf 76.9 & \bf 24.1 & \bf 66.4  & \bf 84.4 & \bf 51.2 & \bf 76.8  \\
        \bottomrule
    \end{tabular}
\end{table*}

In this section, we demonstrate \sys accurately understands users in conversations by evaluating its language understanding capabilities on ground truth data.
Next, we evaluate the effectiveness of \sys for model understanding by performing a real-world human study on healthcare workers (e.g., doctors and nurses) and ML practitioners, where we benchmark \sys against existing explainability systems.
We find users both prefer and are more effective using \sys than traditional point-and-click explainability systems, demonstrating its effectiveness for understanding ML models.

\subsection*{Language Understanding}

Here, we quantitatively assess the language understanding capabilities of \sys by creating gold parse datasets and evaluating the system's accuracy on this data.

\paragraph{Gold Parse Collection} We construct gold datasets (i.e., ground truth (utterance, parse) pairs) across multiple datasets to evaluate the language understanding performance of our models.
To construct these gold datasets, we adopt an approach inspired by~\citet{YuSpider18}, which constructs a similar dataset for multitask semantic parsing.

Our gold dataset generation process is as follows.
First, we write $50$ (utterance, parse) pairs for the particular task (i.e., loan or diabetes prediction).
These utterances range from  simple \userinput{How likely are people in the data to have diabetes?} to complex \userinput{If these people were not unemployed, what's the likelihood they are good credit risk? Why?} and conversational \userinput{What if they were twenty years older?}.
We include each operation (Table~\ref{tab:operations}) at least twice in the parses, to make sure there is good coverage.
From there, we ask Mechanical Turk workers to rewrite the utterances while preserving their semantic meaning to ensure that the ground truth parse for the revised utterance is the same but the phrasing differs.
 We ask workers to rewrite each pair $8$ times for a total of $400$ (utterance, parse) pairs per task. Next, we filter out low-quality mturk revisions.
We ask the crowd sourced workers to rate the similarity between the original utterance and revised utterance on a scale of ($1$-$4$), where $4$ indicates the utterances have the same meaning and $1$ that they do not have the same meaning.
We collect $5$ ratings per revision and remove (utterance, parse) pairs that score below $3.0$ on average.
Finally, we perform an additional filtering step to ensure data quality by inspecting the remaining pairs ourselves and removing any bad revisions.

Since we want to evaluate \sys's capacity to generalize across different scenarios, we perform this data collection process across $3$ different tasks: Pima Indian Diabetes Dataset~\cite{Dua:2019}, German Credit Dataset~\cite{Dua:2019}, and the COMPAS recidivism dataset~\cite{compas}.
After collecting revisions and ensuring quality, we are left with $200$ pairs for German Credit, $190$ for diabetes, and $146$ for COMPAS.

\paragraph{Models} 

We compare two strategies for using pre-trained LLMs to parse user utterances into the grammar \textbf{1.)} few-shot GPT-J \cite{gpt-j} and \textbf{2.)} finetuned T5 \cite{2020t5}.
 Both these models translate user utterances into the \sys grammar in a seq2seq fashion.
However, the GPT-J models are higher-capacity and more amenable to be trained by in-context learning.
This procedure includes examples of the input and target from the training prepended to the test instance~\cite{NEURIPS2020_1457c0d6, min2022rethinking, Xie2021AnEO}.
On the other hand, the T5 models require traditional finetuning on the input and target pairs. 
Consequently, the few-shot approach is quicker to set up because it does not require finetuning, making it easier for users to get started with the system. 
However, the finetuned T5 leads to improved performance and a better user experience overall while taking longer to set up.
To train these models through finetuning or prompting, we generate synthetic (utterance, parse) pairs because it is impractical to assume that we can collect ground truth pairs for every new task we wish to use \sys.
We provide additional training details in the methods section.

We evaluate both fine-tuned T5 models and few-shot GPT-J models on the testing data.
We additionally implement a naive nearest neighbors baseline, where we select the closest user utterance in the synthetic training set according to cosine distance of \texttt{all-mpnet-base-v2} sentence embeddings and return the corresponding parse~\cite{reimers-2019-sentence-bert}.
For the GPT-J models, we compare $N$-Shot performance, where $N$ is the number of (utterance, parse) pairs from the synthetically generated training sets included in the prompt, and sweep over a range of $N$ for each model.
For the larger models, we have to use relatively smaller $N$ in order for inference to fit on a single $48$GB GPU. 

When parsing the utterances, one issue is that their generations are unconstrained and may generate parses outside the grammar, resulting in the system failing to run the parse and bad user experiences.
To ensure the generations are grammatical, we constrain the decodings to be in the grammar \cite{semanticmachines-2021-emnlp}.
This technique, referred to as \textit{guided decoding}, constrains the LLM generations to only allow those tokens that appear next in the grammar at any point during generation.
Practically, we accomplish this by recompiling the grammar at inference time into an equivalent grammar consisting of the tokens in the LLM's vocabulary.
While decoding from the LLM, we fix the likelihood of ungrammatical tokens to $0$ at every generation step. Thus, the LLM only generates grammatical parses.

\paragraph{Evaluating The Parsing Accuracy}
To evaluate performance on the datasets, we use the exact match parsing accuracy~\cite{talmor-etal-2017-evaluating, YuSpider18, gupta22exact}.
This metric is whether the parse exactly matches the gold parse in the dataset.
In addition, we perform the evaluation on two splits of each gold parse dataset, in addition to the overall dataset.
These splits are the IID and compositional splits.
The IID split contains (utterance, parse) pairs where the parse's \parse{operations} and their structure (but not necessarily the arguments) are in the training data.
The compositional split consists of the remaining parses that are not in the training data.
Because LM's struggle compositionally, this split is generally much harder for LM's to parse~\cite{Oren2020ImprovingCG, yin-etal-2021-compositional}.

\paragraph{Accuracy}

We present the results in Table~\ref{tab:parsing_results}.
The fine-tuned T5 performs better overall than the few shot GPT-J models.
In particular, the T5 Large models perform strongly on both the IID and compositional data and can even parse complex compositional phrases.
Notably, the T5 small model performs better than the GPT-J 6B model, which has two orders of magnitude more parameters.
This dynamic is particularly true in the compositional splits in the data where the GPT-J few shot models never exceed $10\%$ parsing accuracy.
Overall, these results indicate \sys can understand user utterances with a high degree of accuracy using our best performing T5 models.
Further, we recommend using this model for the best results and use it for our remaining evaluation.

\subsection*{User Study: Utility of Explainability Dialogues}
The results in the previous subsection show \sys understands user intentions to a high degree of accuracy.
In this subsection, we evaluate how well the end-to-end system helps users understand ML models compared to current explainability systems.

\paragraph{Study Overview} We compare \sys against \textit{explainerdashboard}, one of the most popular open-source explainability dashboards~\cite{oege_dijk_2022_6408776}.
This dashboard has similar functionality to \sys, considering it provides an accessible way to compute explanations and perform model analyses.
Thus, it is a reasonable baseline.
Last, we perform this comparison using the Diabetes dataset, and a gradient boosted tree trained on the data~\cite{scikit-learn}.
To compare both systems in a controlled manner, we ask participants to answer general ML questions with \sys and the dashboard.
Each question is about basic explainability and model analysis, and participants answer using multiple choice, where one of the options is ``Could not determine.'' if they cannot figure out the answer (though it is straightforward to answer all the questions with both interfaces).
For example, questions are about comparing feature importances ``Is glucose more important than age for the model's predictions for data point 49?'' or model predictions ``How many people are predicted not to have diabetes but do not actually have it?''
Participants answer $10$ total questions.
We divide the $10$ questions into $2$ blocks of $5$ questions each.
Both blocks have similar questions but different values to control for memorization (exact questions given in Supplementary Information~\ref{ap:sec:hyperparams}).
Participants use \sys to answer one block of questions and the dashboard for the other block.
In addition, we provide a tutorial on how to use both systems before showing users the questions for the system.
Last, we randomize question, block, and interface order to control for biases due to showing interfaces or questions first.

\begin{table}
    \centering
    \caption{
        User study results: \% of respondents that agree (> Neutral Likert score) \sys is better than the dashboard in the $4$ comparison questions. A significant portion of respondents agreed \sys is better than the dashboard in all the categories except Grad. students and ``Likeliness To Use''. Still, a majority agreed \sys was superior in this case.\vspace{1mm}
    }
    \label{tab:percieved-results}
    \begin{tabular}{lcc}
        \toprule
        & \multicolumn{2}{c}{\% Agree \sys Better} \\ 
        \cmidrule(lr){2-3}
         Comparison &  \Centerstack[c]{Health Care Workers} & \Centerstack[c]{ML Grad. Students} \\
        \midrule
         Easiness & 82.2 & 84.6 \\
         Confidence & 77.7 & 69.2 \\
         Speed & 84.4 & 84.6 \\
         Likeliness To Use & 73.3 & 53.8 \\
        \bottomrule
    \end{tabular}
\end{table}

\paragraph{Metrics} 
Following previous work on evaluating human and ML coordination and trust, we assess several metrics to evaluate user experiences~\cite{chen2022hint, radar2006, glassTowardsTrust2008}.
We evaluate the following statements along $1$-$7$ Likert scale at the end of the survey:
\begin{itemize}
    \item \textbf{Easiness:} \textit{I found the conversational interface easier to use than the dashboard interface}
    \item \textbf{Confidence:} \textit{I was more confident in my answers using the conversational interface than the dashboard interface}
    \item \textbf{Speed:} \textit{I felt that I was able to more rapidly arrive at an answer using the conversational interface than the dashboard interface}
    \item \textbf{Likeliness To Use:} \textit{Based on my experience so far with both interfaces, I would be more likely to use the conversational interface than the dashboard interface in the future}
\end{itemize}
To control for bias associated with the ordering of the terms conversational interface and dashboard interface, we randomized their ordering.
We also measure accuracy and time-taken to answer each question.
Last, we asked to participants to write a short description comparing their experience with both interfaces to capture participants qualitative feedback about both systems.

\paragraph{Recruitment} Since \sys provides an accessible way to understand ML models, we expect it to be useful for subject matter experts with a variety of experience in ML, including users without any ML experience.
As such, we recruited $45$ English speaking healthcare workers to take the survey using the Prolific service~\cite{prolific} with minimal or no ML expertise
This group comprises a range of healthcare workers, including doctors, pharmacists, dentists, psychiatrists, healthcare project managers, and medical scribes.
The vast majority of this group ($43$) stated they had either no experience with ML or had heard about it from reading articles online, while two members indicated they had equivalent to an undergraduate course in ML.
As another point of comparison, we recruited ML professionals with relatively higher ML expertise from ML Slack channels and email lists.
We received $13$ potential participants, all of which had graduate course level ML experience or higher, and included all of them in the study.
We received IRB approval for this study from our institution's IRB approval process and informed consent from participants.

\paragraph{Metric Results} A significant majority of health care workers agreed they preferred \sys in all the categories we evaluated (Table~\ref{tab:percieved-results}).
The same is true for the ML professionals, save for whether they were more likely to use \sys in the future, where 53.8\% of participants agreed they would instead use \sys in the future.
In addition, participants subjective notions around how quickly they could use \sys aligned with their actual speed of use, and both groups arrived at answers using \sys significantly quicker than the dashboard.
The median question answer time (measured at the total time taken from seeing the question to submitting the answer) using \sys was $76.3$ seconds, while it was $158.8$ seconds using the dashboard. 

Participants were also much more accurate and completed questions at a higher rate (i.e., they did not mark ``could not determine) using \sys (Table \ref{tab:accuracy-results}).
While both health care workers and ML practitioners clicked could not determine for a quarter of the questions using the dashboard, this was true for $13.8\%$ of health care workers and $6.1\%$ of ML professionals using \sys, demonstrating the usefulness of the conversational interface.
On completed questions, both groups were much more accurate using \sys than the dashboard.
Most surprisingly, though ML professionals agreed they preferred \sys only about half the time, they answered all the questions correctly using it, while they only answered $62.5\%$ of questions correctly with the dashboard.
Finally, we observed \sys's conversational capabilities were highly effective.
There were only $6$ utterances out of over $1,000$ total utterances the conversational aspect of the system failed to resolve. 
These failure cases generally involved certain discourse aspects like asking for additional elaboration (``more description'').

The largest source of errors for participants using the explainability dashboard were two questions concerning the top most important features for individual predictions.
The errors for these questions account for $47.4\%$ of health care workers and $44.4\%$ of ML professionals' total mistakes.
Answering these questions with the dashboard requires users to perform multiple steps, including choosing the feature importance tab in the dashboard, selecting local explanations for the correct instance, and ranking the features according to their importance. 
On the other hand, the streamlined text interface of TalkToModel made it much simpler to solve these questions resulting fewer errors.

\begin{table}
    \centering
    \caption{
        User study results: Completion rate and accuracy across interfaces and participant groups. We compute the completion rate as the questions users provided and answer for and did not mark ``could not determine.'' We measure accuracy on completed questions. Participants answered questions at a higher rate more accurately using \sys than the dashboard. \vspace{1mm}
    }
    \label{tab:accuracy-results}
    \begin{tabular}{lcccc}
        \toprule
        & \multicolumn{2}{c}{\Centerstack[c]{\% Questions \\Completed}} & \multicolumn{2}{c}{\Centerstack[c]{\% Accuracy On\\Completed Questions}} \\ 
        \cmidrule(lr){2-3} \cmidrule(lr){4-5} 
        &  Dash. & \sys & Dash. & \sys \\
        \midrule
        \vspace{1mm}\Centerstack[l]{Health Care Workers} & $74.7$ & $86.2$ & $66.1$ & $91.8$ \\
        \Centerstack[l]{ML Grad. Students} & $73.8$ & $93.9$ & $62.5$ & $100.0$ \\ 
        \bottomrule
    \end{tabular}
\end{table}

\paragraph{Qualitative Results} 
For the qualitative user feedback, we provide representative quotes from similar themes in the responses.
Users expressed that they could more rapidly and easily arrive at results, which could be helpful for their professions,
\begin{displayquote}
\textit{I prefer the conversational interface because it helps arrive at the answer very quickly. This is very useful especially in the hospital setting where you have hundreds of patients getting check ups and screenings for diabetes because it is efficient and you can work with medical students on using the system to help patient outcomes.}—P39 medical worker at a tertiary hospital.
\end{displayquote}
Participants also commented on the user friendliness of \sys and its strong conversational capabilities, stating, ``the conversational [interface] was straight to the point, way easier to use''—P35 Nurse, and that ``the conversational interface is hands-down much easier to use... it feels like one is talking to a human.''—P45 ML Professional.
We did not find any negative feedback surrounding the conversational capabilities of the system.
Users also commented on how easy it was to access information compared to the dashboard,
\begin{displayquote}
\textit{With the conversational interface you can ask whatever you want to know and with the dashboard you need to specifically search information that you don't actually know where it is.}—P31 Physical Therapist.

\end{displayquote}
All in all, users expressed strong positive sentiment about \sys due to the quality of conversations, presentation of information, accessibility, and speed of use.

Several ML professionals brought up points that could serve as future research directions.
Notably, participants stated they would rather look at the data themselves rather than rely on an interface that rapidly provides an answer,
\begin{displayquote}
\textit{I would almost always rather look at the data myself and come to a conclusion than getting an answer within seconds.}—P11 ML Professional.
\end{displayquote}
In the future, it would be worthwhile including visualizations of raw data and analyses performed by the system to increase trust with expert users, such as ML professionals, who may be skeptical of the high-level answers provided by the system currently.

\section{Discussion}

With ML models increasingly becoming more complex, there is need to develop techniques to explain model predictions to stakeholders.
Nevertheless, it is often the case that practitioners struggle to use explanations and often have many follow up questions they wish to answer.
In this work, we show \sys makes explainable AI accessible to users that come from a range of backgrounds by using natural language conversations.
Our experimental findings demonstrate \sys both comprehends users to a high-degree of accuracy and can help users understand the predictions of ML models much better than existing systems.
In particular, we showed \sys is a highly effective way for domain experts such as healthcare workers to understand ML models, like those applied to disease diagnosis. 
Last, we designed \sys to be highly extensible and release the code, data, and a demo for the system at \texttt{\href{https://github.com/dylan-slack/TalkToModel}{https://github.com/dylan-slack/TalkToModel}}, making it straightforward for explainability users and researchers to build on the system.

In the future, it will be helpful to investigate applications of \sys \textit{in-the-wild}, such as in doctors' offices, laboratories, or professional settings, where model stakeholders use the system to understand their models.
In addition, it will also be helpful to explore how to use language models to generate conversation responses grounded in the results of the operations.
Finally, it will also be helpful to evaluate how best to integrate \sys in existing scientific and professional work streams to promote its impact and usefulness.

\section{Methods}
\label{sec:methods}
\dylan{Note methods \& supp. info do NOT count in the word count of 3500}

In this section, we describe the components of \sys. 
First, we introduce the dialogue engine and discuss how it understands user inputs, maps them to operations, and generates text responses based on the results of running the operations.
Second, we describe the execution engine, which runs the operations.
Finally, we provide an overview of the interface and the extensibility of \sys.

\subsection{Text Understanding}

To understand the intent behind user utterances, the system learns to translate or \textit{parse} them into logical forms.
These parses represent the intentions behind user utterances in a highly-expressive and structured programming language \sys executes.

Compared to dialogue systems that execute specific tasks by modifying representations of the internal state of the conversation~\cite{hongshenchen2017, li2017end-to-end}, our parsing-based approach allows for more flexibility in the conversations, supporting open-ended discovery, which is critical for model understanding.
Also, this strategy produces a structured representation of user utterances instead of open-ended systems that generate unstructured free text~\cite{Santhanam2019ASO}.
Having this structured representation of user inputs is key for our setting where we need to execute specific operations depending on the user's input, which would not be straightforward with unstructured text.

\sys performs the following steps to accomplish this: 
\textbf{1)} the system constructs a grammar for the user-provided dataset and model, which defines the set of acceptable parses,
\textbf{2)} \sys generates (utterance, parse) pairs for the dataset and model,
\textbf{3)} the system finetunes a large language model (LLM) to translate user utterances into parses, and \textbf{4)}
the system responds conversationally to users by composing the results of the executed parse into a response that provides context for the results and opportunities to follow up.

\paragraph{Grammar} 

To represent the intentions behind the user utterances in a structured form, \sys relies on a grammar, defining a domain specific language for model understanding.
While the user utterances themselves will be highly diverse, the grammar creates a way to express user utterances in a structured yet highly expressive fashion that the system can reliably execute.
Compared with approaches that treat determining user intentions in conversations as a classification problem~\cite{liu-etal-2017-using-context, caipredicting2020}, using a grammar enables the system to express compositions of operations and arguments that take on many different values, such as real numbers, that would otherwise be combinatorially impossible in a prediction setting.
Instead, \sys translates user utterances into this grammar in a seq2seq fashion, overcoming these challenges~\cite{sutskever2014}.
This grammar consists of production rules that include the \parse{operations} the system can run (an overview is provided in Table~\ref{tab:operations}), the acceptable \args{arguments} for each operation, and the relations between operations.
One complication is that user-provided datasets have different feature names and values, making it hard to define one shared grammar between datasets.
Instead, we update the grammar based on the feature names and values in a new dataset.
For instance, if a dataset only contained the feature names \texttt{age} and \texttt{income}, these two names would be the only acceptable values for the \args{feature} argument in the grammar.

To ensure our grammar provides sufficient coverage for XAI questions, we very our grammar supports the questions from the \textit{XAI question bank}.
This question bank was introduced by~\citet{Liao2020QuestioningTA} based on interviews with AI product designers and includes $31$ core, prototypical questions XAI systems should answer, excluding socio-technical questions beyond the scope of \sys (e.g., What are the results of other people using the [model]).
The prototypical questions address topics such as the input/data to the model (\userinput{What is the distribution of a given feature?}), model output (\userinput{What kind of output does the system give?}), model performance (\userinput{How accurate are the predictions?}), global model behavior (\userinput{What is the systems overall logic?}), why/why not the system makes individual predictions (\userinput{Why is this instance given this prediction?}), and what-if or counterfactual questions (\userinput{What would the system predict if this instance changes to...?}).
To evaluate how well \sys covers these questions, we review each question and evaluate whether our grammar can parse it.
Overall, we find our grammar supports $30/31$ of the prototypical questions. 
We provide a table of each question and corresponding parse in Supplementary Table~\ref{tab:xai-question-bank1} and Supplementary Table~\ref{tab:xai-question-bank2}.
Overall, the grammar covers the vast majority of XAI related questions, and therefore, has good coverage of XAI topics.

\paragraph{Supporting Context In Dialogues}
User conversations with \sys naturally include complex conversational phenomena such as anaphora and ellipsis~\cite{grosz-etal-1983-providing, tseng-etal-2021-cread, NEURIPS2018_d63fbf8c}. Meaning, conversations refer back to events earlier in the conversation (\userinput{what do you predict for \textit{them}?}) or omit information that must be inferred from conversation (\userinput{Now show me for people predicted incorrectly.}). 
However, current language models only parse a single input, making it hard to apply them in settings where the context is important.
To support context in the dialogues, \sys introduces on a set of operations in the grammar that determine the context for user utterances.
In contrast with approaches that maintain the conversation state using neural representations~\cite{hongshenchen2017, gao-etal-2019-dialog}, grammar operations allow for much more trustworthy and dependable behavior while still fostering rich interactions, which is critical for high-stakes settings, and similar mechanisms for incorporating grammar predicates across turns have been shown to achieve strong results~\cite{NEURIPS2018_d63fbf8c}.
In particular, we leverage two operations: \parse{previous filter} and \parse{previous operation}, which look back in the conversation to find the last filter and last operation, respectively.
These operations also act recursively.
Therefore, if the last filter is a \parse{previous filter} operation, \sys will recursively call \parse{previous filter} to resolve the entire stack of filters.
As a result, \sys is capable of addressing instances of anaphora and ellipsis by using these operations to resolve the entity via co-reference or infer it from the previous conversation history.
This dynamic enables users to have complex and natural conversations with \sys.

\paragraph{Parsing Dataset Generation} 
To parse user utterances into the grammar, we finetune an LLM to translate utterances into the grammar in a seq2seq fashion.
We use LLMs because these models have been trained on large amounts of text data and are solid priors for language understanding tasks.
Thus, they can better understand diverse user inputs than training from scratch, improving the user experience.
Further, we automate the finetuning of an LLM to parse user utterances into the grammar by generating a training dataset of (utterance, parse) pairs.
Compared to dataset generation methods that use human annotators to generate and label datasets for training conversation models~\cite{gao-etal-2018-neural-approaches, Rieser2012}, this approach is much less costly and time consuming, while still being highly effective, and supports users getting conversations running very quickly.
This strategy consists of writing an initial set of user utterances and parses,
where parts of the utterances and parses are \textit{wildcard} terms.
\sys enumerates the wildcards with aspects of a user-provided dataset, such as the feature names, to generate a training dataset.
Depending on the user-provided dataset schema, \sys typically generates anywhere from $20,000$-$40,000$ pairs. 
Last, we have already written the initial set of utterances and parses, so users only need to provide their dataset to setup a conversation.

\paragraph{Semantic Parsing} Here, we provide additional details about the semantic parsing approach for translating user utterances into the grammar.
The two strategies for parsing user utterances using pre-trained LLMs that we considered were \textbf{1.)} few-shot GPT-J \cite{gpt-j} and \textbf{2.)} finetuned T5 \cite{2020t5}.
With respect to the few-shot models, because the LLM's context window only accepts a fixed number of inputs, we introduce a technique to select the set of most relevant prompts for the user utterance.
In particular, we embed all the utterances and identify the closest utterances to the user utterance according to the cosine distance of these embeddings.
To ensure a diverse set of prompts, we only select one prompt per template.
We prompt the LLM using these (utterance, parse) pairs, ordering the closest pairs immediately before the user utterance because LLMs exhibit recency biases~\cite{Zhao2021Calibrate}.
Using this strategy, we experiment with the number of prompts included in the LLM's context window.
In practice, we use the \texttt{all-mpnet-base-v2} sentence transformer model to perform the embeddings~\cite{reimers-2019-sentence-bert}, and we consider the GPT-J 6B, GPT-Neo 2.7B, and GPT-Neo 1.3B models in our experiments.

We also fine-tune pre-trained T5 models in a seq2seq fashion on our datasets.
To perform fine-tuning, we split the dataset using a 90/10\% train/validation split and train for $20$ epochs to maximize the next token likelihood with a batch size of $32$. 
We select the model with the lowest validation loss at the end of each epoch.
We fine-tune with a learning rate of $1e$-$4$ and the AdamW optimizer \cite{Loshchilov2019DecoupledWD}.
Last, our experiments consider the T5 Small, Base, and Large variants.

\paragraph{Generating Responses}
After \sys executes a parse, it composes the results of the operations into a natural language response it returns to the user.
\sys generates these responses by filling in templates associated with each operation based on the results.
The responses also include sufficient context to understand the results and opportunities for following up (examples in Table~\ref{tab:conversation}).
Further, because the system runs multiple operations in one execution, \sys joins response templates, ensuring semantic coherence, into a final response and shows it to the user.
Compared to approaches that generate responses using neural methods~\cite{Shao2017GeneratingHA}, this approach ensures the responses are trustworthy and do not contain useless information hallucinated by the system, which would be a very poor user experience for the high-stakes applications we consider. Further, because \sys supports a wide variety of different operations, this approach ensures sufficient diversity in responses, so they are not repetitive.

\subsection{Executing Parses}
\label{subsec:executing}

In this subsection, we provide an overview of the execution engine, which runs the operations necessary to respond to user utterances in the conversation.
Further, this component automatically selects the most faithful explanations for the user, helping ensure explanation accuracy.

\paragraph{Feature Importance Explanations}
At its core, \sys explains why the model makes predictions to users with feature importance explanations.
Feature importance explanations $\phe ( \x,\, \f ) \rightarrow \imp $ accept a data point $\x \in \mathbb{R}^d$ with $d$ features and model as input $\f ( \x ) \rightarrow \y$, where $\y \in [0, 1]$ is the probability for a particular class, and generates a feature attribution vector $\imp \in \mathbb{R}^d$, where greater magnitudes correspond to higher importance features \cite{slackreliable21, smilkov2017smoothgrad, lime:whi16, yeh2019fidelity, chen2018lshapley, pmlr-v139-agarwal21c}.

We implement the feature importance explanations using \textit{post-hoc} feature importance explanations.
Post-hoc feature importance explanations do not rely on internal details of the model $\f$ (e.g., internal weights or gradients) and only on the input data $\x$ and predictions $\y$ to compute explanations, so users are not limited to only certain types of models~\cite{ribeiro2016should, Lundberg2020fromlocal, faithfulLakk2019, maplePlumb2018, learningTheoretic}.
Note that our system can easily be extended to other explanations that rely on internal model details, if required~\cite{gradcam, definitionsmethodsMurdoch2019, NEURIPS2021_251bd044, sundararajanAxiomaticAttribution2017}.

\paragraph{Explanation Selection} 
While there exists several post hoc explanation methods, each one adopts a different definition of what constitutes an explanation~\cite{krish2022disagree}.
For example, while LIME, SHAP, and Integrated Gradients all output feature attributions, LIME returns coefficients of local linear models, SHAP computes Shapley values, and Integrated Gradients leverages model gradients.
Consequently, we automatically select the \textit{most faithful} explanation for users, unless a user specifically requests a certain technique.
Following prior works, we compute faithfulness by perturbing the most important features and evaluating how much the prediction changes~\cite{Meng2022}.
Intuitively, if the feature importance $\phi$ correctly captures the feature importance ranking, perturbing more important features should lead to greater effects.

While previous works \cite{Lundberg2020fromlocal, roarHooker}, compute the faithfulness over many different thresholds, making comparisons harder, or require retraining entirely from scratch, we introduce a single metric that captures the prediction sensitivity to perturbing certain features called the \textit{fudge score}.
This metric is the mean absolute difference between the model's prediction on the original input and a fudged version on $\textbf{m} \in \{ 0, 1 \}^d$ features,
\begin{align}
\textrm{Fudge}(\f, \x, \textbf{m}) = \frac{1}{N} \sum_{n=1}^{N} \; \lvert \f( \x ) - \f ( \x + \eps_n \odot \textbf{m} ) \rvert
\end{align}
where $\odot$ is the tensor product and $\eps\sim\mathcal{N}(0,I\sigma)$ is $N \times d$ dimensional Gaussian noise.
To evaluate faithfulness for a particular explanation method, we compute area under the fudge score curve on the top-k most important features, thereby summarizing the results into a single metric,
\begin{align}
     \topkimpindices ( k, \imp ) & = \begin{cases} 1 & \textrm{if  } \phi_i \in \argmax_{\imp'\subset \{1..d\}, |\imp'|=k} \sum_{i \in \imp'} | \phi_i |  \\
     0 & \textrm{otherwise}
     \end{cases} 
     \\
    \textrm{Faith}& (\imp,\, f,\, \x,\, K) = \sum_{k=1}^{K}  \textrm{Fudge} (f,\, \x,\, \topkimpindices ( k, \imp ))
    \label{eq:auc}
\end{align}
where $\topkimpindices(k, \imp)$ is the indicator function for the top-k most important features.
Intuitively, if a set of feature importances $\imp$ correctly identifies the most important features, perturbing them will have greater effects on the model's predictions, resulting in higher faithfulness scores.
We compute faithfulness for multiple different explanations and select the highest.
In practice, we consider LIME~\cite{ribeiro2016should} with the following kernel widths $[0.25, 0.50, 0.75, 1.0]$ and KernelSHAP~\cite{lundberg2017unified}. We leave all settings to default besides the kernel widths for LIME.
In practice, we set $\sigma=0.05$ to ensure perturbations happen in the local region around the prediction, $K$ to $\textrm{floor}(\frac{d}{2})$, and $N=10,000$ to sample sufficiently.
One complication arises for categorical features, where we cannot apply Gaussian perturbations.
For these features, we randomly sample these features from a value in the dataset column $30\%$ of the time to guarantee the feature remains categorical under perturbation.
Last, if multiple explanations return similar fidelities, we use the explanation stability metric proposed by~\citet{AlvarezMelis2018OnTR} to break ties, because it is much more desirable for the explanation to robust to perturbations~\cite{slackreliable21, rethinkingStability}. 
In order to use the \textit{stability} metric proposed by \citet{AlvarezMelis2018OnTR} to break ties if the explanations fidelities are quite close (less than $\delta=0.01$), we compute the jaccard similarity between feature rankings instead of the $l2$ norm as is used in their work.
The reason is that the norm might not be comparable between explanation types, because they have different ranges, while the jaccard similarity should not be affected.
Further, we compute the area under the top k curve using the jaccard similarity stability metric, as in Equation~\ref{eq:auc}, to make this measure more robust.

\paragraph{Additional Explanation Types}
Since users will have explainability questions that cannot be answered solely with feature importance explanations, we include additional explanations to support a wider array of conversation topics.
In particular, we support counterfactual explanations and feature interaction effects.
These methods enable conversations about \textit{how} to get different outcomes 
and if features \textit{interact} with each other during predictions, supporting a broad set of user queries.
We implement counterfactual explanations using DiCE, which generates a diverse set of counterfactuals \cite{mothilal2020dice}.
Having access to many plausible counterfactuals is desirable because it enables users to see a breadth of different, potentially useful, options.
Also, we implement feature interaction effects using the partial dependence based approach from~\citet{Greenwell2018ASA} because it is effective and quick to compute.

\paragraph{Exploring Data and Predictions}
Because the process of understanding models often requires users to inspect the model's predictions, errors, and the data itself, \sys supports a wide variety of data and model exploration tools.
For example, \sys provides options for filtering data and performing what-if analyses, supporting user queries that concern subsets of data or what would happen if data points change.
Users can also inspect model errors, predictions, prediction probabilities, compute summary statistics, and evaluation metrics for individuals and groups of instances.
\sys additionally supports summarizing common patterns in mistakes on groups of instances by training a shallow decision tree on the model errors in the group.
Also, \sys enables descriptive operations, which explain how the system works, summarize the dataset, and define terms to help users understand how to approach the conversation.
Overall, \sys supports a rich set of conversation topics in addition to explanations, making the system a complete solution for the model understanding requirements of end users.

\subsection*{Extensibility}
While we implement \sys with several different choices for operations such as feature importance explanations and counterfactual explanations, \sys is highly modular and system designers can easily incorporate new operations or change existing ones by modifying the grammar to best support their user populations.
This design makes \sys straightforward to extend to new settings, where different operations may be  desired.

\section*{Data Availability} 
The German, Compas, and Diabetes datasets and models can be found at \texttt{\href{https://github.com/dylan-slack/TalkToModel/tree/main/data}{https://github.com/ dylan-slack/TalkToModel/tree/main/data}}. The finetuned language models used for TalkToModel for each of these datasets can be found at \texttt{\href{https://huggingface.co/dslack/all-finetuned-ttm-models}{https://huggingface.co/dslack/all-finetuned-ttm-models}}.
The mturk generated dataset used to assess parsing accuracy and the accuracy results can be found at \texttt{\href{https://github.com/dylan-slack/TalkToModel/tree/main/experiments/parsing_accuracy}{https://github.com/dylan-slack/TalkToModel/tree/main/experiments/parsing\_accuracy}}. The user study response data is provided at \texttt{\href{https://github.com/dylan-slack/TalkToModel/blob/main/data/ttm-user-study-responses.csv}{https://github.com/dylan-slack/TalkToModel/blob/main/ data/ttm-user-study-responses.csv}}.

\section*{Code Availability}
We release an open source implementation of \sys at \texttt{\href{https://github.com/dylan-slack/TalkToModel}{https://github.com/dylan-slack/ TalkToModel}}. The DOI is \texttt{\href{https://doi.org/10.5281/zenodo.7502206
}{https://doi.org/10.5281/zenodo.7502206}}.
Beyond the methods described so far, this release includes visualizations for conversations, interactive tooling to help users construct questions, saving results and conversation environments so they can be shared, abstractions for creating new operations and synthetic datasets, routines to adapt \sys to new models and datasets automatically, and runtime optimizations (generating responses typically takes $<2$ seconds).

\nocite{slidesgo}
\bibliography{pnas-sample}

\section{Broader Impact Statement}

The \sys system and, more generally, conversational model explainability can be applied to a wide range of applications, including financial, medical, or legal applications.
Our research could be used to improve model understanding in these situations by improving transparency and encouraging the positive impact of ML systems, while reducing errors and bias.
Although \sys has many positive applications, the system makes it easier for those without high levels of technical expertise to understand ML models, which could lead to a false sense of trust in ML systems.
In addition, because \sys makes it easier to use ML model for those with lower levels of expertise, there is additionally a risk of inexperienced users applying ML models inappropriately.
While \sys includes several measures to prevent such risks, such as qualifying when explanations or predictions are inaccurate, and clearly describing the intended purpose of the ML model, it would be useful for researchers to investigate and possible adopters to be mindful of these considerations.

\section{Acknowledgements}

The authors would like to acknowledge helpful feedback from Peter Hase, Johan Ugander, Marco Tulio Ribeiro, Brian Lim, and the UCI NLP lab concerning earlier versions of the system and manuscript.
This work is supported in part by the NSF awards \#IIS-2008461,  \#IIS-2040989, \#IIS-2046873, \#IIS-2008956, and research awards from Google, JP Morgan, Amazon, Harvard Data Science Initiative, the $D^3$ Institute at Harvard, and the Hasso Plattner Institute. H.L. would like to thank Sujatha and Mohan Lakkaraju for their continued support and encouragement. 
The views expressed here are those of the authors and do not reflect the official policy or position of the funding agencies.

\section{Contributions}

D.S. designed and developed the \sys system, studies, and prepared the manuscript.
S.K. designed and implemented the explanation selection procedure, drafted sections in the manuscript, and editied the manuscript.
S.S. and H.L. contributed equally to advising the development of the system and experiments, conceiving the system, reviewing the manuscript, and editing the manuscript.

\section{Competing Interests}

The authors do not declare any competing interests.

\clearpage
\onecolumn
\appendix
\begin{center}
    \LARGE
    \textbf{Supplementary Information}
\end{center}
\renewcommand{\tablename}{Supplementary Table}
\renewcommand{\figurename}{Supplementary Figure}
\setcounter{figure}{0}
\setcounter{table}{0}
\setcounter{equation}{0}

\section{Experimental Details}
\label{ap:sec:hyperparams}

In this appendix, we provide an overview of additional experimental details.
First, we provide additional details about the grammar.
Next, we describe how we generate our parsing training sets.
Further, we provide more details about the explanation selection procedure.
After, we provide examples of the sentences we had mturk workers revise to create our gold parsing datsets.
Finally, we provide additional details about the user study.

\subsection{Grammar Details}

In this subsection, we provide additional details about the grammar. 
First, we describe the design of the grammar.
After, we provide details about how we update the grammar for new datasets.

\paragraph{Design}
Recall from the main paper that the grammar serves as a logical form of user utterances, which \sys can execute.
Here, we provide more details about the grammar design.
The grammar defines relations between the \parse{operations} and the acceptable values \args{arguments} in Table~\ref{tab:operations}.
For example, the grammar defines different acceptable values for the \args{comparison} argument in the \parse{filter} operation, such as \args{less than or equal to} or \args{greater than}. In addition, we structure the grammar to make parses appear closer to natural language text instead of a formal programming language like SQL or Python. 
The reason is that language models tend to perform better at translating utterances into grammars that are more similar to natural language instead of a programming language~\cite{semanticmachines-2021-emnlp}.
Consequently, we design the grammar, so that parses appear more like natural language text, without unnecessary parentheses and commas, and omitting unnecessary arguments where possible.
In general, we found that simplifying the grammar and making it more like natural language as much as possible considerably improved performance.
For example, the question \userinput{What are the three most important features for people older than thirty-five?} would simply translate to \parse{filter} \args{age greater than 35} \parse{and topk} \args{3}.
Note, here, because \sys is applied to only one dataset at a time, the \args{dataset} argument can be omitted for simplicity.
Practically, we implement the grammar in Lark~\cite{lark} because the implementation supports interactive parsing, simplifying the process of implementing the guided decoding strategy.

\paragraph{Updating the Grammar For Datasets and User Utterances}

In the main paper, we discussed how we update the grammar based on the dataset.
Here, we provide more description about how we update the grammar.
We update the grammar based on the feature names and categorical feature values in the dataset.
In particular, the acceptable values for the \args{feature} argument (Table~\ref{tab:operations}) in the grammar becomes all the feature values in the dataset.
Further, the \args{value} argument for a categorical \args{feature} becomes the set of categorical feature values that appear in the data for that feature.
Because there are many possible values of numeric features, we instead extract potential numeric values from user utterances as they are provided to the system.
Specifically, we set the \args{value} argument in the grammar for numeric features to contain the set of numeric values that appear in the user utterance.
We additionally support string based numeric values (e.g., ``fifty-five'' or ``twelve'') to make the system handle a wider variety of cases.

\subsection{Training Dataset Generation}

In this subsection, we provide details about the generation of the (utterance, parse) pair training dataset.
To ensure that we generate a diverse and comprehensive set of (utterance, parse) pairs for training, we compose a total of $687$ templates that use $6$ different wildcard types.
The templates consist of a diverse set of utterances that encompass the different operations permitted in the system.
The wildcards include categorical feature names, numeric feature names, class names, numeric feature values, categorical feature values, explanation types,  and common filtering expressions (e.g., \userinput{\wildcard{NUMERIC\_FEATURE} above \wildcard{NUMERIC\_VALUE}}).
Because templates can have potentially many wildcards, we recursively enumerate all the wildcard values for each parse.
Further, we also limit the number of values for certain wildcards  to ensure the number of training pairs generated does not become extremely large.
In particular, we limit the number of numeric values to $2$ values per feature.
In addition, to prevent templates with many wildcards from dominating the training dataset, we also downsample the number of values per wildcard to $2$ values after the initial recursion.
In this way, we the training dataset does not get dominated by a few templates that have many wildcards, which we found improves performance.

As an example, an utterance template is \userinput{Explain the predictions on data with \wildcard{NUMERIC\_FEATURE} greater than \wildcard{NUMERIC\_VALUE}} and the corresponding parse template is \parse{filter \wildcard{NUMERIC\_FEATURE}}\args{ greater than }\parse{\wildcard{NUMERIC\_VALUE} and explain }\args{feature importance}.
From there, we enumerate the numeric features in the dataset and a selection of numeric feature values, substituting these into \wildcard{NUMERIC\_FEATURE} and \wildcard{NUMERIC\_VALUE} respectively to generate data.

\subsection{Gold Parsing Dataset}
\label{ap:sec:grammar}

In this subsection, we provide examples of $10$ of the sentences provided to mturk workers to revise during our data generation process for each dataset.
The examples are provided in Table~\ref{tab:example_sentences} and illustrate the different types of utterances revised by mturkers to create a comprehensive testing set. Also, we provide the number of (utterance, parse) pairs in the IID and compositional splits in Table~\ref{tab:splits}.

\begin{table}[ht]
    \tiny
    \centering
    \caption{Examples of $10$ sentences out of $50$ from each dataset provided to mturk workers to revise to generate parsing evaluation data.\vspace{1mm}}
    \label{tab:example_sentences}
    \begin{tabular}{ll}
        \toprule
        \multirow{10}{*}{\centering \rotatebox[origin=c]{90}{COMPAS}} & What is your reasoning for determining if people older than 20 are likely to commit crimes? \\
         & How likely are people that are younger than 25 or have committed at least 1 crime in the past to commit a crime in future? \\
         & what are top 3 most important features you use for prediction for people if they were to decrease their prison terms by 10 months? \\
         & For people that are 18 years old and black, how often are you correct in predicting whether they will commit crimes in the future? \\
         & let's look at those in the data with 3 or more prior crimes on record. what are some common mistakes the model makes on these people? \\
         & For this subset in the data, how accurate is the model? \\
         & For people that are 18 years old and black, how often are you correct in predicting whether they will commit crimes in the future? \\
         & how likely would the person with the id number of 33 in the data be to a commit a crime if they were 5 years younger? \\
         & But what if they were twenty years older? \\
         & Could you show me some data for people who are black? \\
         \midrule
         \multirow{10}{*}{\centering \rotatebox[origin=c]{90}{Diabetes}} & How likely are people that either (1) have had two pregnancies or (2) are older than 20 and younger than 30 to have diabetes? \\
         & What are the top five most important features for the model's predictions on people with a bmi over 40? \\
         & Show the data for people older than 20. Then, could you show me the predictions on this data? \\
         & what would happen to the likelihood of having diabetes if we were to increase glucose by 100 for the data point with id 33 \\
         & What's the average age in the data? \\
         & For this subset in the data, how accurate is the model? \\
         & What does patient number 34 need to do in order to be diagnosed as unlikely to have diabetes? \\
         & What are the reasons why the model predicted data point number 100 and what could you do to change it? \\
         &  How would the predictions change if age were decreased by 5 years for people with a bmi of 30? \\
         & How do the features of the data interact in the model's predictions on this particular data? \\
         \midrule
         \multirow{10}{*}{\centering \rotatebox[origin=c]{90}{German}}
         & If people in the data were unemployed, how important would the age and loan amount features be for predicting credit risk? \\
         & what would happen to the likelihood of being bad credit risk if we were to increase the loan amount by 250 for the data point with id 89 \\
         & what are top three most important features for determining whether those who are applying for furniture loans are good credit risk? \\
         & What is the average loan amount for people with no current loans and that do not own a house? \\
         & How accurate are you at predicting whether people who are asking for loans for home appliances are good credit risk? \\
         & In the dataset, if the loan duration were to be increased by 2 years, what would the predictions for the data be? \\
         & Could you show me some examples the model predicts incorrectly and how accurate the model is on the data? \\
         & What do you predict on the instances in the data? Also, could you show me an example of a few mistakes you make in these predictions? \\
         & But why did you think these people are bad credit risk? \\
         & If these people were not unemployed, what's the likelihood they are good credit risk? Why? \\
         \bottomrule
    \end{tabular}
\end{table}

\begin{table}
    \centering
    \caption{The number of gold (utterance, parse) pairs in the IID and Compositional splits for the datasets. There are relatively more IID questions in each dataset.\vspace{1mm}}
    \label{tab:splits}
    \begin{tabular}{lccc}
        \toprule
        & COMPAS & Diabetes & German \\
        \midrule
        IID & $117$ & $147$ & $127$ \\
        Compositional & $29$ & $43$ & $74$ \\
        Overall & $146$ & $190$ & $201$ \\
        \bottomrule
    \end{tabular}
\end{table}

\subsection{User Study Questions}

In this subsection, we provide the questions participants answered in the user study. The questions are provided in Table~\ref{tab:question-bank}.
One of the two question blocks, Block 1 or Block 2, is shown for \sys and the other is shown for the dashboard (the ordering of \sys and the dashboard is also randomized).
The two question blocks include similar concepts to ensure a similar level of difficulty but include different numbers to discourage memorization between the question blocks.

\begin{table}[ht]
\tiny
    \centering
        \caption{The tasks solved by participants during the user study. Participants were shown either one block for \sys and the other block for the dashboard.\vspace{1mm}}
    \label{tab:question-bank}
    \begin{tabular}{ll}
        \toprule
         \multirow{5}{*}{\centering \rotatebox[origin=c]{90}{Block 1}} & What are the three most important features for the model's predictions on people older than 30 with bmi's above 35?  \\
        & What is the feature importance ranking of the age feature for data point id 188? \\
        & How many individuals in the dataset are predicted to be likely to have diabetes but are not actually likely to have it? \\
        & If patient id 293 were to decrease their bmi by five, what's the prediction probability of the "likely to have diabetes" class? \\
        & Is the "glucose" feature more important than the "age" feature for the model's prediction on data point 49? \\
        \midrule
        \multirow{5}{*}{\centering \rotatebox[origin=c]{90}{Block 2}} & What are the three most important features for the model's predictions on people younger than 23 with glucose levels below seventy-five? \\
        & What is the feature importance ranking of the insulin feature for data point id 57? \\
        & How many individuals in the dataset are predicted not likely to have diabetes but actually are likely to have it? \\
        & What's the likelihood of patient 57 having diabetes if they increased their glucose levels by 100 and bmi by 3? \\
        & How important is the "diabetes pedigree function" feature compared to the "glucose" feature for the model's prediction on data point 55? \\
        \bottomrule
    \end{tabular}
\end{table}

\subsection{User Study Length \& Compensation}

The survey took around $30$ minutes for participants to complete.
We compensated the ML professionals by providing them with a $\$20$ gift card.
We paid the prolific workers $\$14.74$/hour on average for completing the survey, considering their individual completion times.

\section{Additional Experiments}
\label{ap:sec:adexperiments}

In this appendix, we provide additional experimental results.
First, we provide experimental results where we show the advantages of explanation selection.
Second, we give results on the affects of the number of training templates.
After, we provide error analysis for our parsing models.
Last, we give additional user study results.

\subsection{Advantages of Explanation Selection}

In the main paper, we introduced a technique for explanation selection, which we used in \sys to automatically select high-quality explanations for the conversation (Subsection~\ref{subsec:executing}).
In this appendix, we provide more details about the advantages of explanation selection.
In particular, we rigorously benchmark our selection method against SOTA explanation techniques like LIME and SHAP. We show that our explanation selection method computes the most faithful explanations. 

To perform this analysis, we use the faithfulness metrics provided by the widely-used OpenXAI framework~\cite{agarwal2022openxai}.
Specifically, we use the Prediction Gap on Important feature perturbation (PGI) and the Prediction Gap on Unimportant feature perturbation (PGU) metrics.
These metrics measure the change in perturbing the most influential features and least important features, respectively.
Intuitively, PGI captures that perturbations to influential features should result in more significant changes to predictions (higher PGI is better). PGU captures that perturbations to non-influential features should result in smaller changes to the prediction (lower PGU is better).

We compare our explanation selection method against both SOTA explanation methods LIME and SHAP~\cite{lime, lundberg2017unified}. 
To make our evaluation more comprehensive, we compare against LIME using $4$ different settings of the kernel width hyperparameter $[0.25, 0.50, 0.75, 1.0]$, because this hyperparameter can have significant effects on the resulting explanation.
We leave all settings to default otherwise.
Further, we perform this comparison using our $3$ diverse datasets: Diabetes, COMPAS, and German Credit, and we compute explanations three times for each data point to reduce error due to explanation sampling.
We set the important features used for the PGI metric to the most influential 50\% of features and the unimportant features used for the PGU metric to the least influential 50\% of features to ensure we provide a comprehensive evaluation for the explanation's ranking of all features in the data.

We present the results in Table~\ref{tab:fidelity-benchmark} and provide the mean PGU or PGI value for each explanation and dataset.
Further, we bold the best statistically significant result according to a Bonferroni corrected t-test, where we compare the explanation selection procedure to each of the other explanation methods for the respective dataset and metric.
Overall, we find that explanation selection performs better than baseline SOTA explanations across almost every dataset and metric considered, except for the PGU metric on the German dataset, where explanation selection performs on par with the best performing LIME explanations.

\begin{table}
    \centering
    \caption{The prediction gap on important features (PGI) and prediction gap on unimportant features (PGU) results. We bold the statistically significant best result. Overall, explanation selection is the best explanation method in all settings, except for PGU and the german credit data where it is better than SHAP but not significantly better than LIME.\vspace{1mm}}
    \label{tab:fidelity-benchmark}
    \begin{tabular}{lcccccc}
        \toprule
        & \multicolumn{3}{c}{\Centerstack[c]{PGI$\,\uparrow$}} & \multicolumn{3}{c}{\Centerstack[c]{PGU$\,\downarrow$}} \\
        \cmidrule(lr){2-4}  \cmidrule(lr){5-7}
        & Diabetes & COMPAS & German & Diabetes & COMPAS & German \\ \midrule
        LIME, Width=$0.25$ & $0.070$ & $0.124$ & $3.897$ & $0.032$ & $0.031$ & $\boldsymbol{0.774}$  \\
        LIME, Width=$0.50$ & $0.072$ & $0.127$ & $3.871$ & $0.020$ & $0.027$ & $\boldsymbol{0.793}$ \\
        LIME, Width=$0.75$ & $0.071$ & $0.127$ & $3.856$ & $0.021$ & $0.026$ & $\boldsymbol{0.808}$ \\
        LIME, Width=$1.0$ & $0.070$ & $0.127$ & $3.853$ & $0.022$ & $0.026$ & $\boldsymbol{0.799}$ \\
        SHAP & $0.083$ & $0.117$ & $2.094$ & $0.031$ & $0.031$ & $3.007$ \\
        Explanation Selection & $\boldsymbol{0.107}$ & $\boldsymbol{0.138}$ & $\boldsymbol{4.011}$ & $\boldsymbol{0.006}$ & $\boldsymbol{0.023}$ & $\boldsymbol{0.788}$ \\
        \bottomrule
    \end{tabular}
\end{table}

\subsection{Effects of the number of training templates}

\begin{figure}[ht]
    \centering
    \includegraphics[width=0.5\columnwidth]{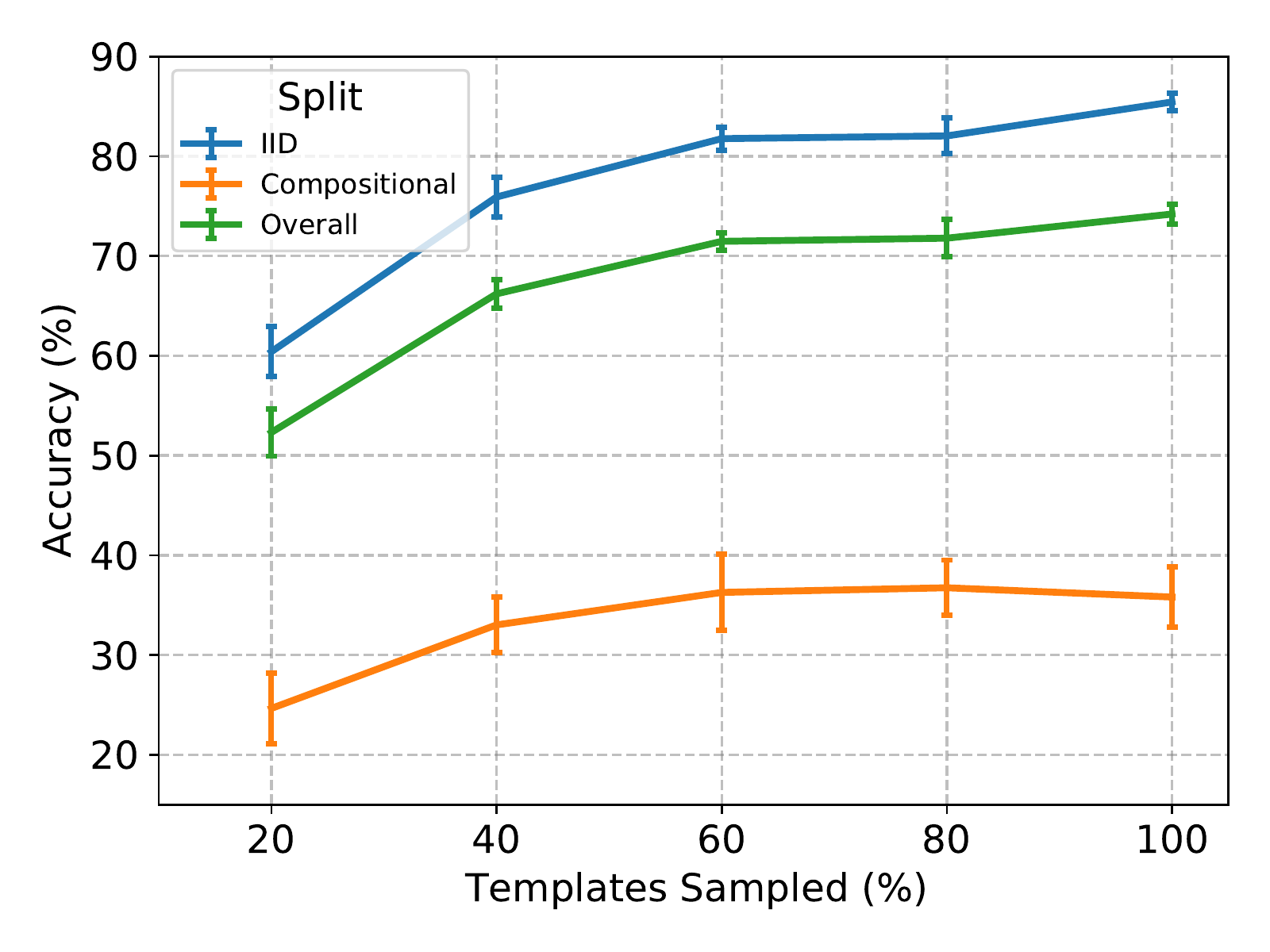}
    \caption{Randomly sampling prompt templates and re-training T5-Base on the Diabetes dataset. For each down-sample \%, the prompts are randomly down-sampled and the model is re-trained $5$ times. The error bars are $1$ standard deviation.}
    \label{fig:diabtes-t5-base-sweep}
\end{figure}

In this subsection, we provide experimental details about the effects of the number of training templates on the parsing accuracy of the system.
Because we use a template strategy to generate training data (Section~\ref{sec:methods}), we must decide on how many prompts to write and include in the training scheme.
This raises the question of how the number of templates affects model performance.
To understand this behavior, we retrain the T5-Base model, randomly sampling the number of training templates over different percentages of the original templates set.
In particular, we sweep over the following percentages $[20\%, 40\%, 60\%, 80\%, 100\%]$ on the diabetes dataset, downsampling and retraining $5$ times per template. We give the results in Figure~\ref{fig:diabtes-t5-base-sweep}.
We see that there are clear accuracy gains over using a smaller number of templates.
Further, the gains for the compositional model performance seem to somewhat level off, but this is not the case for the IID split, suggesting further templates may help IID performance.

\subsection{Parsing Error Analysis}
\begin{table}
    \centering
    \begin{tabular}{lcccc}
        \toprule
        & \multicolumn{2}{c}{5-Shot} & \multicolumn{2}{c}{10-Shot} \\
        \cmidrule(lr){2-3} \cmidrule(lr){4-5}
        & IID & Compositional & IID & Compositional  \\
        \midrule
        Compas   & 29.8 & 77.8  & 15.3 & 53.6 \\
        Diabetes & 40.0 & 40.0 & 19.1 & 28.2  \\
        German   & 44.2 & 50.8 & 28.3 & 26.9 \\
        \bottomrule
    \end{tabular}
    \caption{
        {Percentage of mistakes for few-shot GPT-J 6B where selected prompts \textit{do not} include all the operations in the parse of the user utterance. We see that most of the time the operations for the parse of the user utterance are included in the prompts for the 10-Shot models, yet these methods still perform relatively poorly compared to finetuned T5.}
    }
    \label{tab:failures}
\end{table}

In the main text, we demonstrated that the fine-tuned T5 models performed considerably better than few-shot GPT-J (Table~\ref{tab:parsing_results}).
In this subsection, we perform additional error analysis for why this occurs.
This dynamic brings up the question: what is the cause of these poor few-shot results?
Since the few-shot GPT-J models select the (utterance, parse) pairs from the synthetic dataset using nearest neighbors on a sentence embedding model, it could be possible these issues are due to the sentence embedding model failing to select good pairs.
In particular, this nearest neighbors technique could fail to select pairs with the operations necessary to parse the user utterance, and not the model failing to learn from the examples.
To evaluate whether this is the case, we compute the percent of mistakes that \textit{do not} include the operations necessary to parse the user utterance for GPT-J.
The results provided in Table~\ref{tab:failures} demonstrate that, especially for the 10-Shot case, the operations needed to parse the user utterance \textit{are} included in the prompts, indicating the issues are likely due to the model's capacity to learn few-shot, rather than the selection mechanism.
In this work, we were limited by using up to $6$-billion parameter GPT-J, but it could be possible to achieve better results with larger models, as results on emergent abilitiees suggest~\cite{emergent}.

\subsection{User Study Results: Per Question Likert Scores}

\begin{figure}
    \centering
    \includegraphics[width=.49\columnwidth]{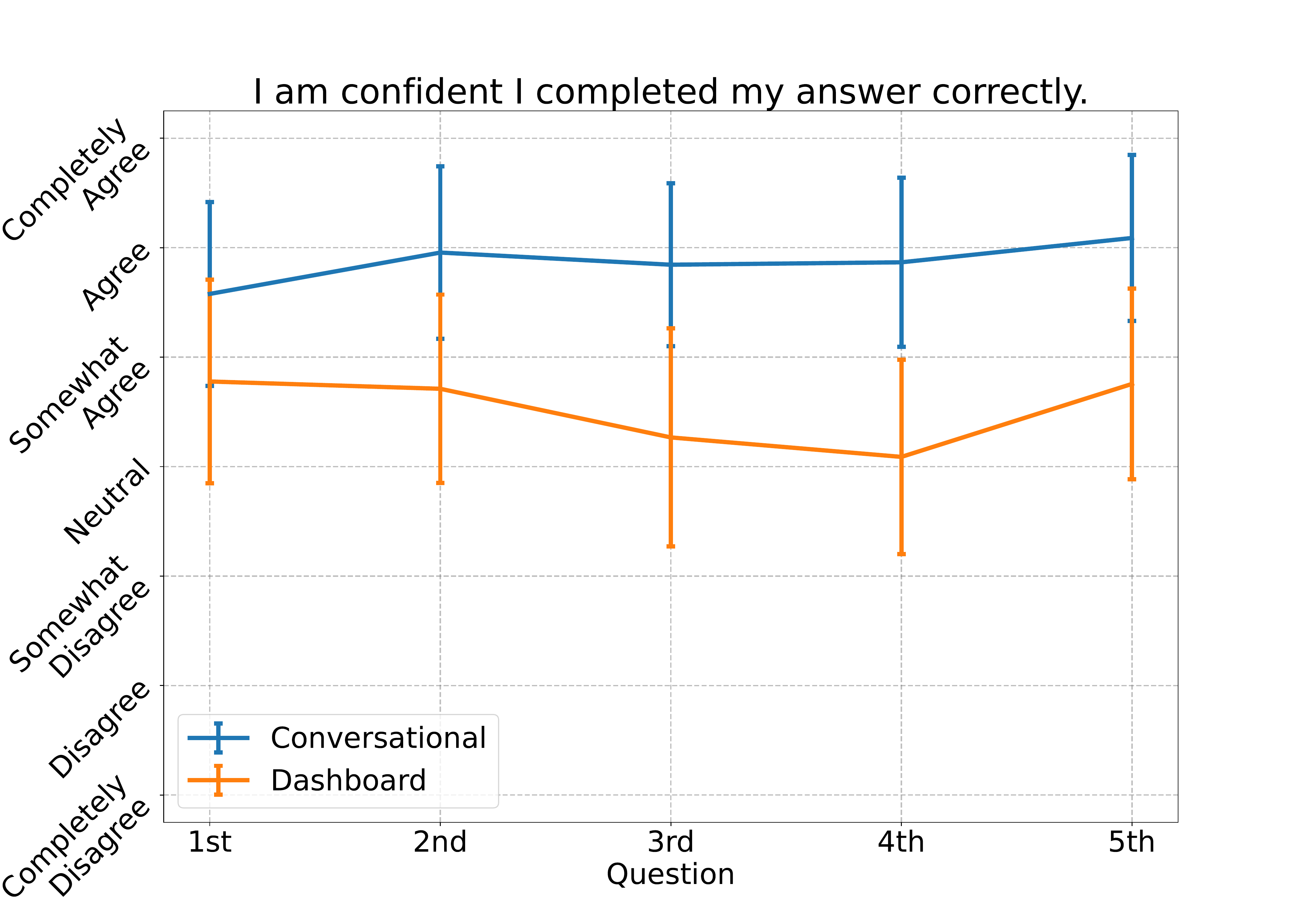}
    \includegraphics[width=.49\columnwidth]{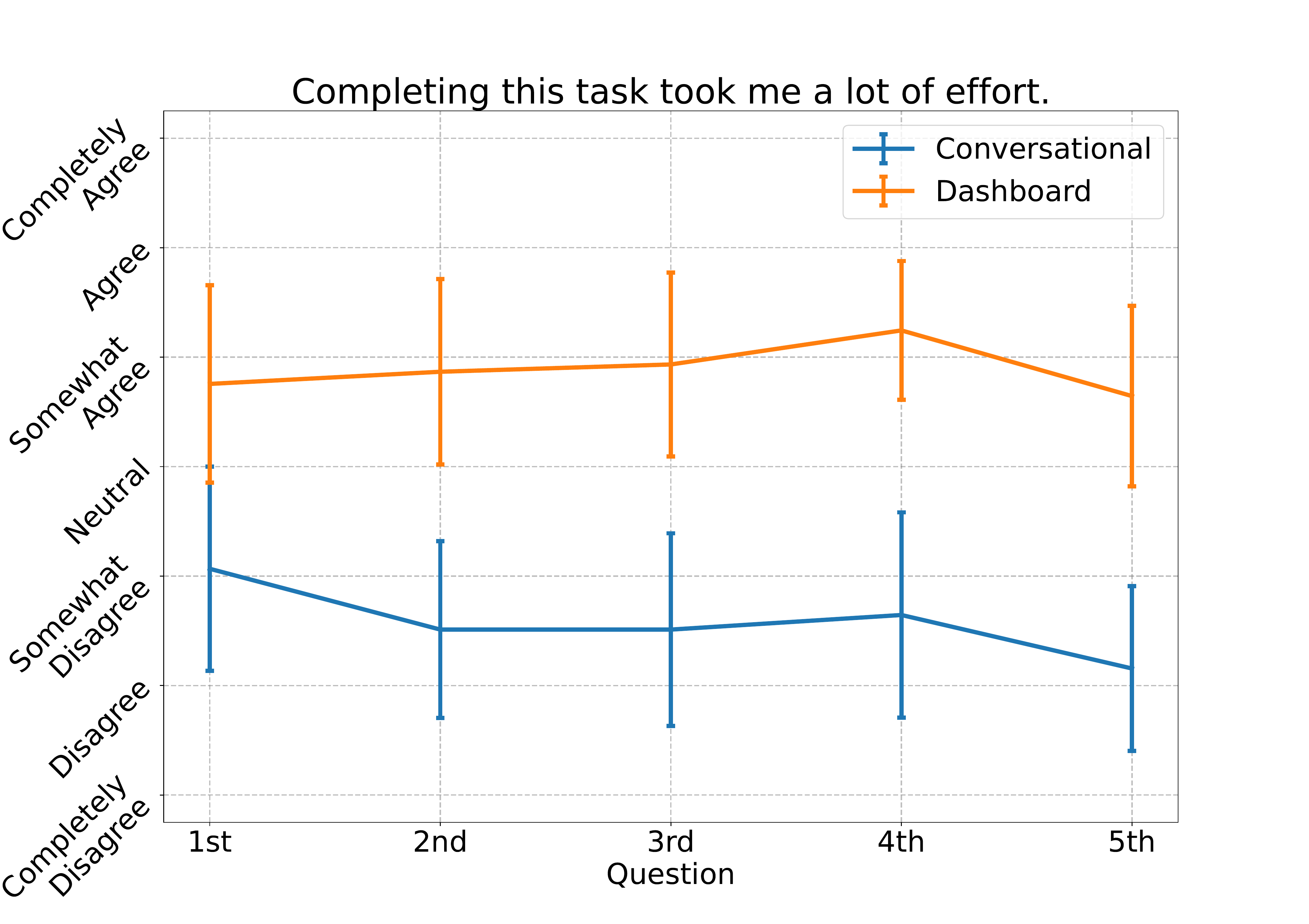}
    \includegraphics[width=.49\columnwidth]{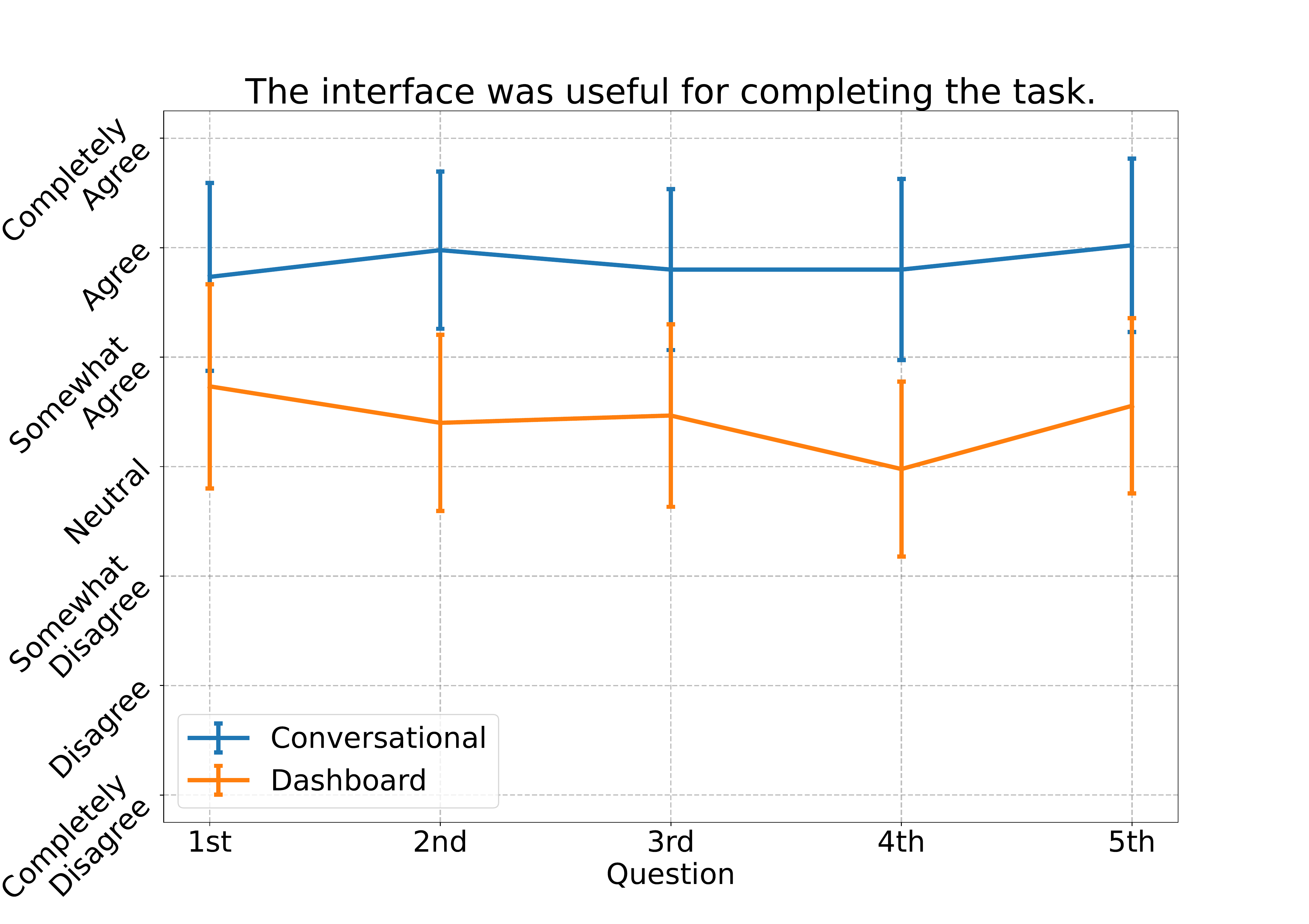}
    \includegraphics[width=.49\columnwidth]{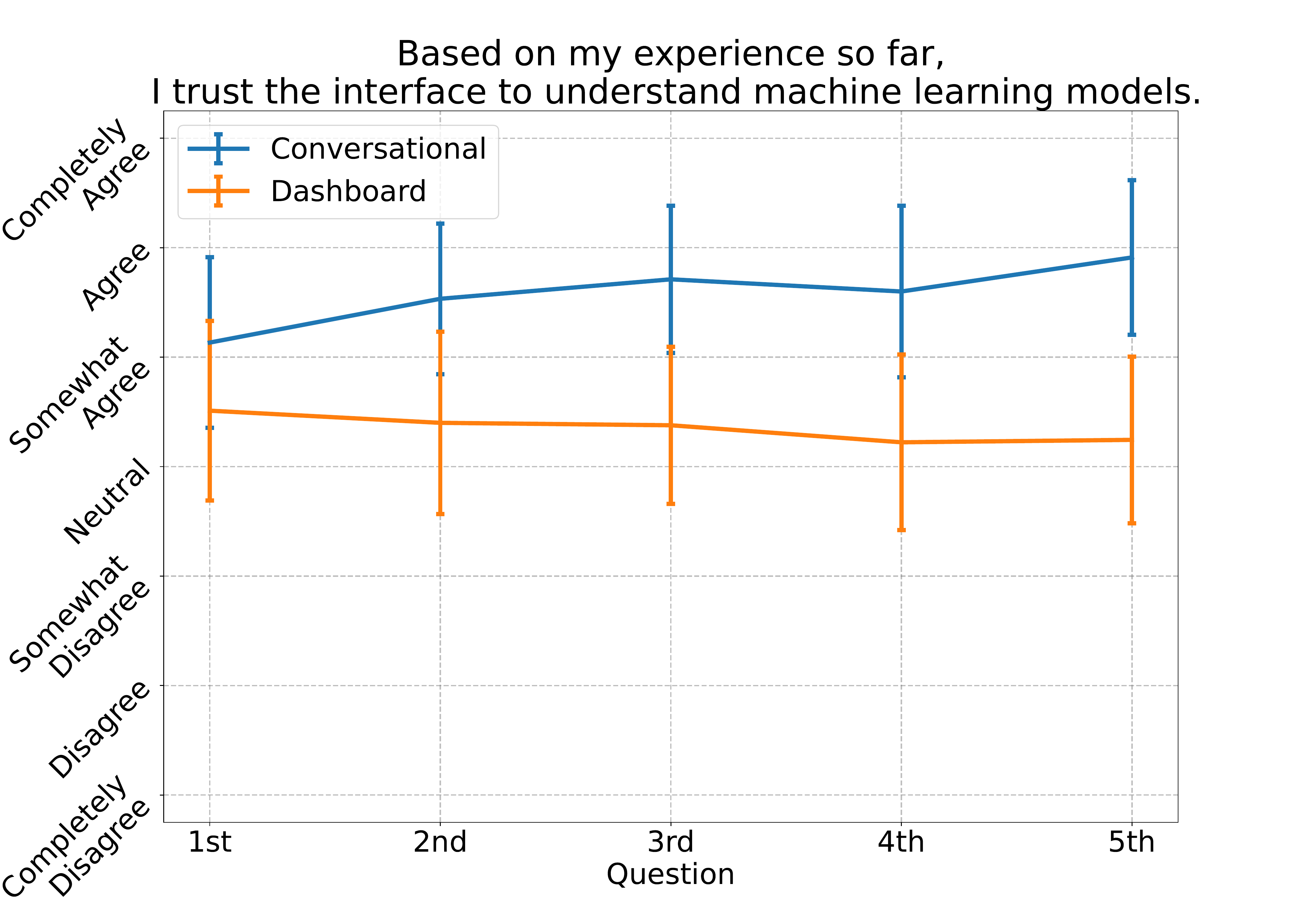}
    \includegraphics[width=.49\columnwidth]{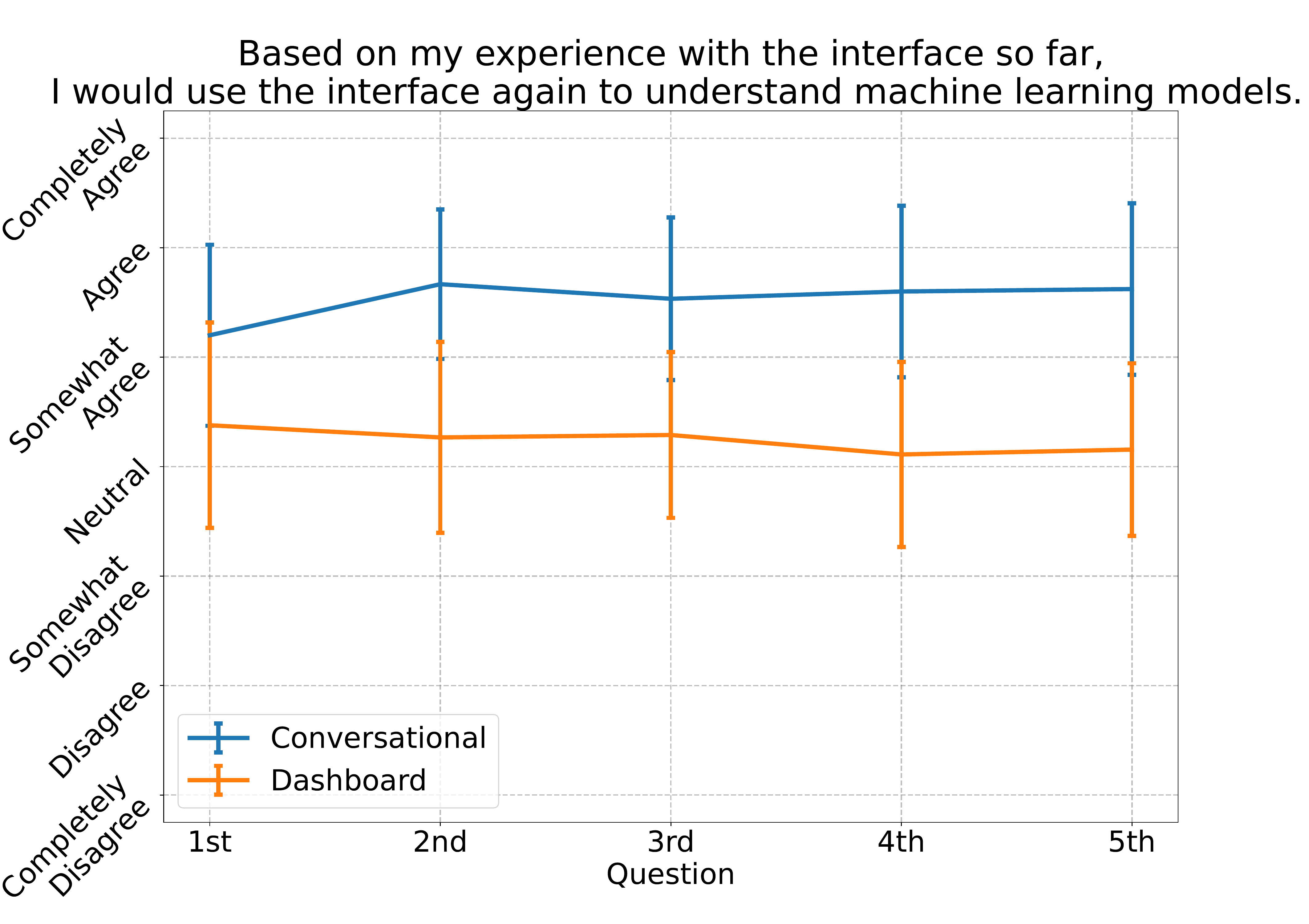}
    \caption{\textbf{Medical Worker per question likert results:} in general, these participants preferred \sys over the dashboard to answer the questions. Error bars are 1 standard deviation.}
    \label{fig:medical-per-q}
\end{figure}

\begin{figure}
    \centering
    \includegraphics[width=.49\columnwidth]{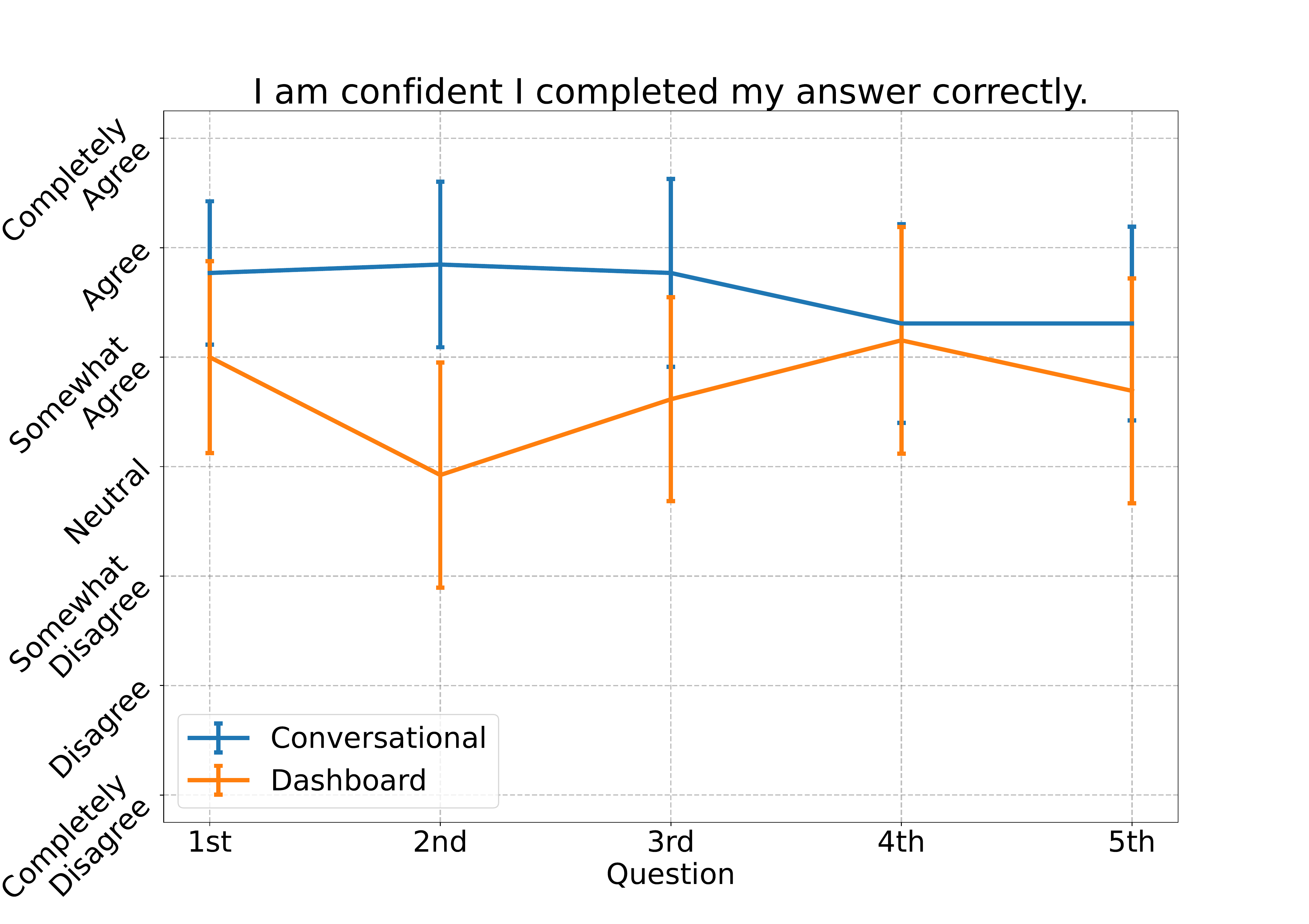}
    \includegraphics[width=.49\columnwidth]{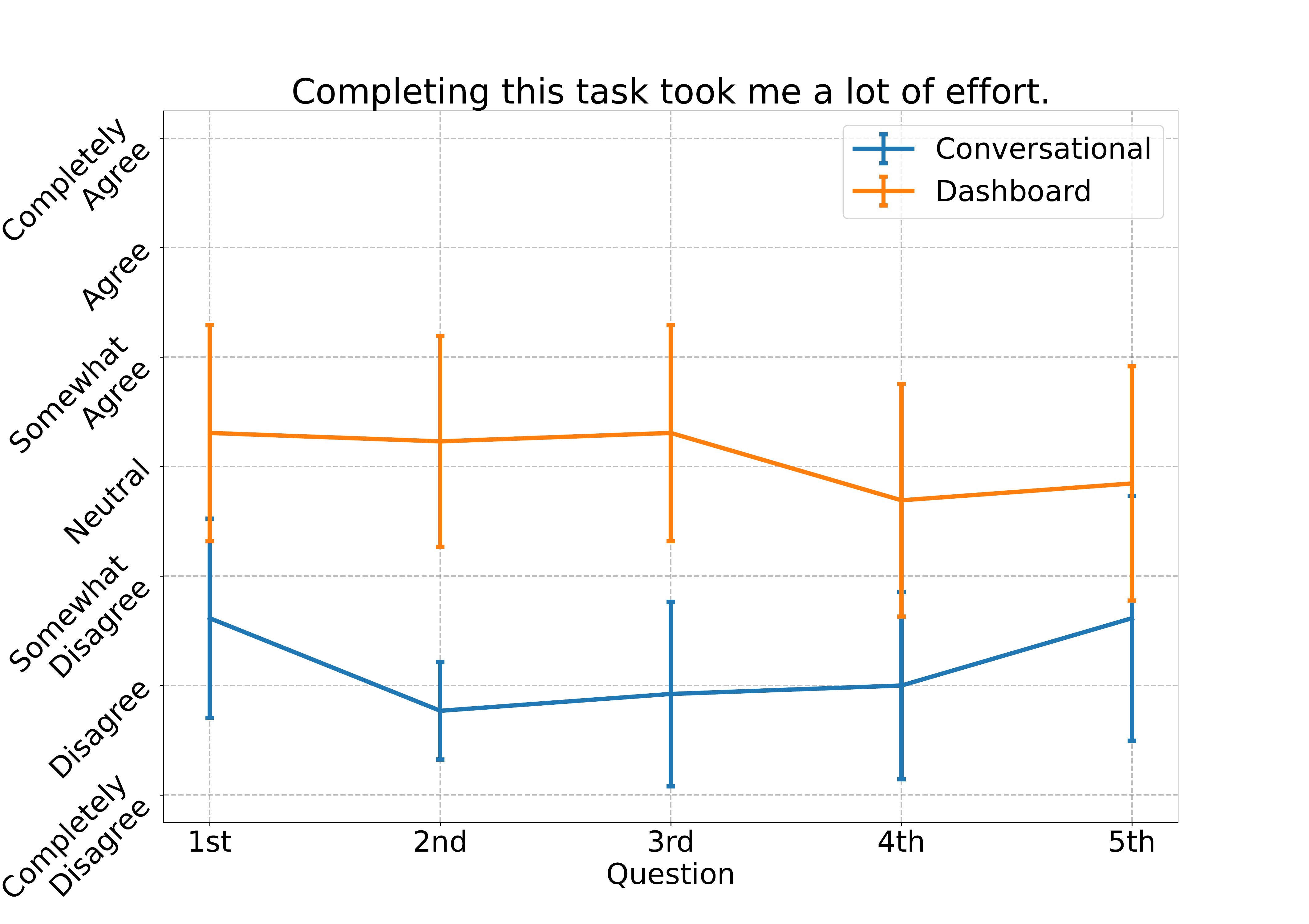}
    \includegraphics[width=.49\columnwidth]{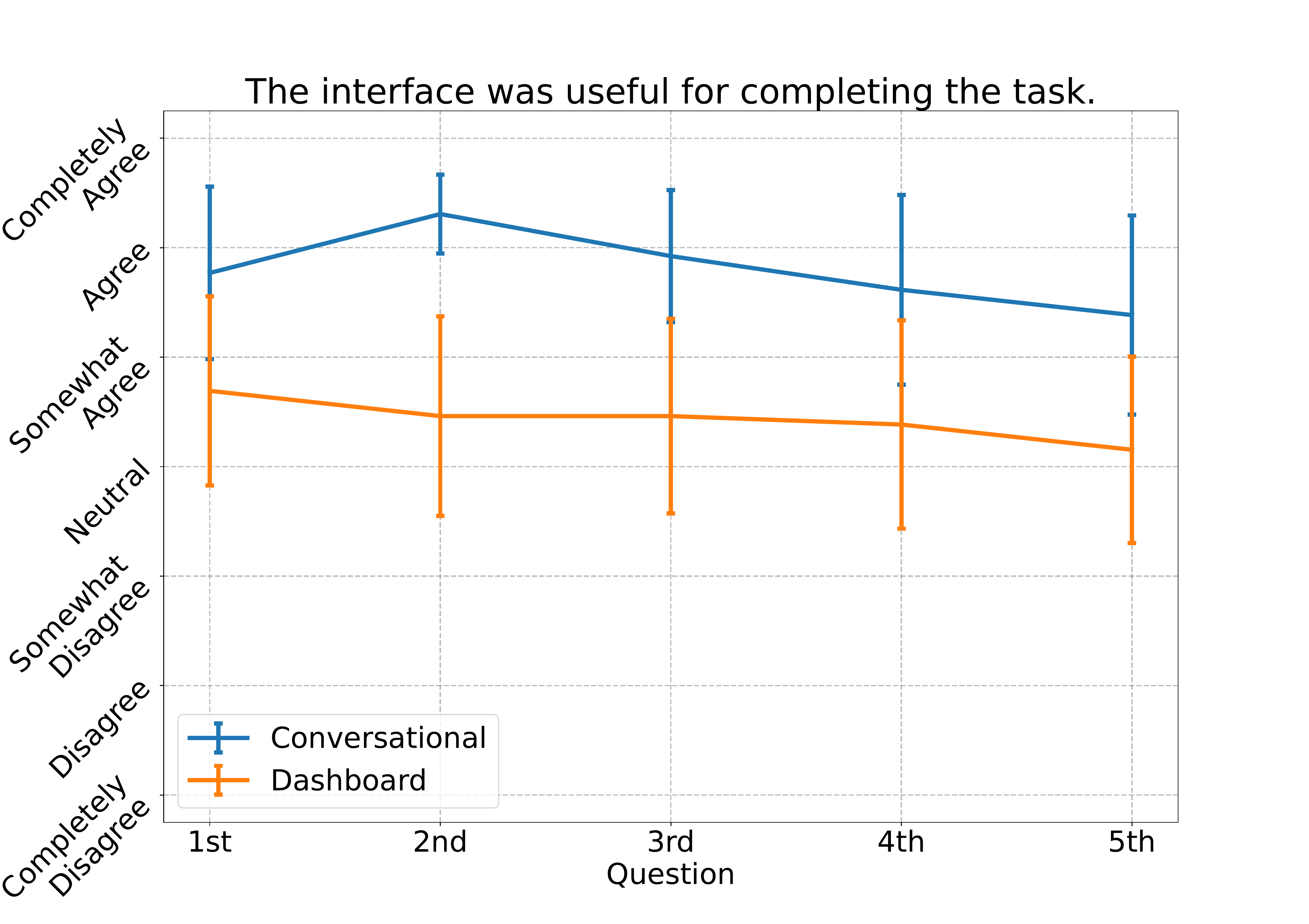}
    \includegraphics[width=.49\columnwidth]{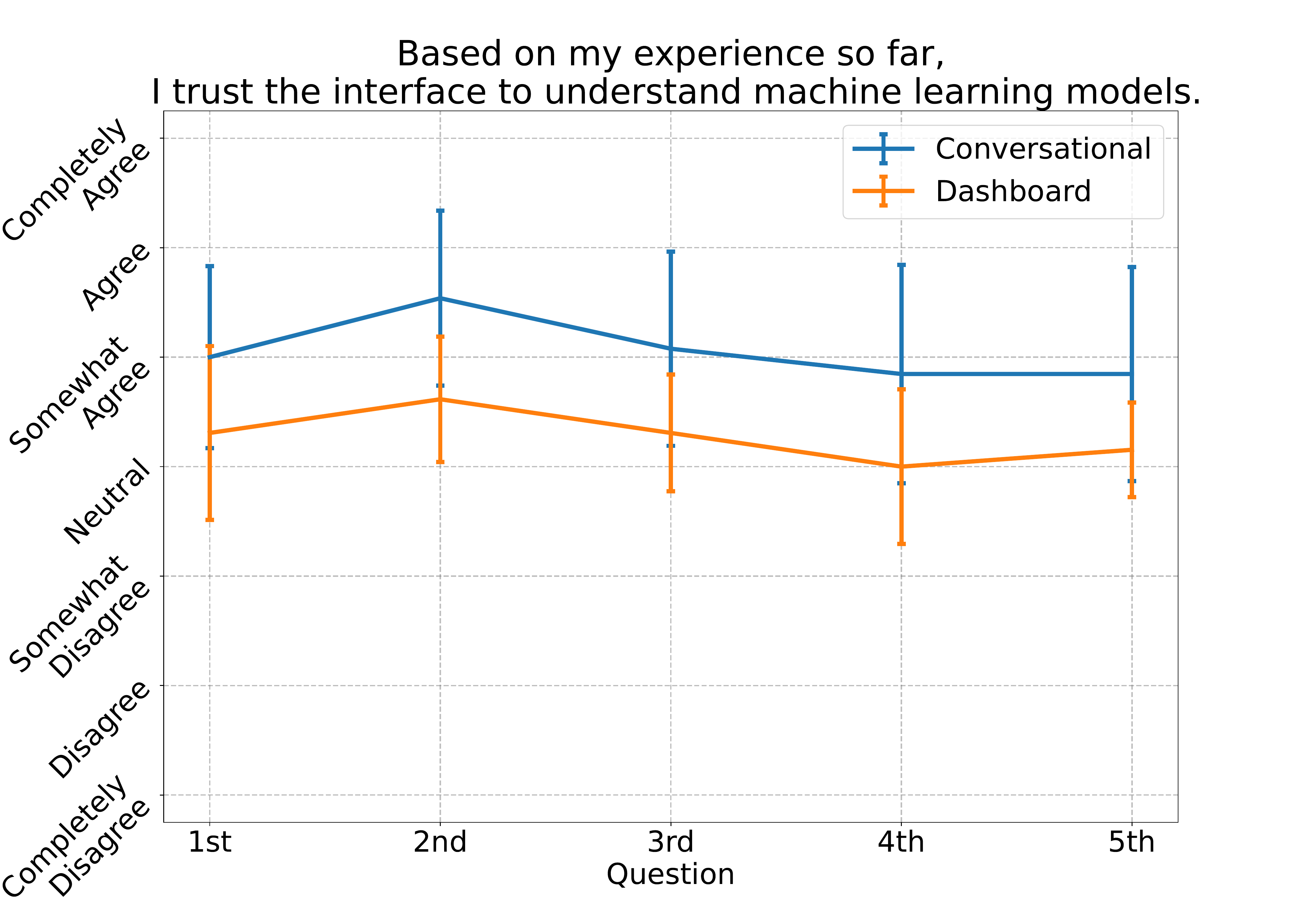}
    \includegraphics[width=.49\columnwidth]{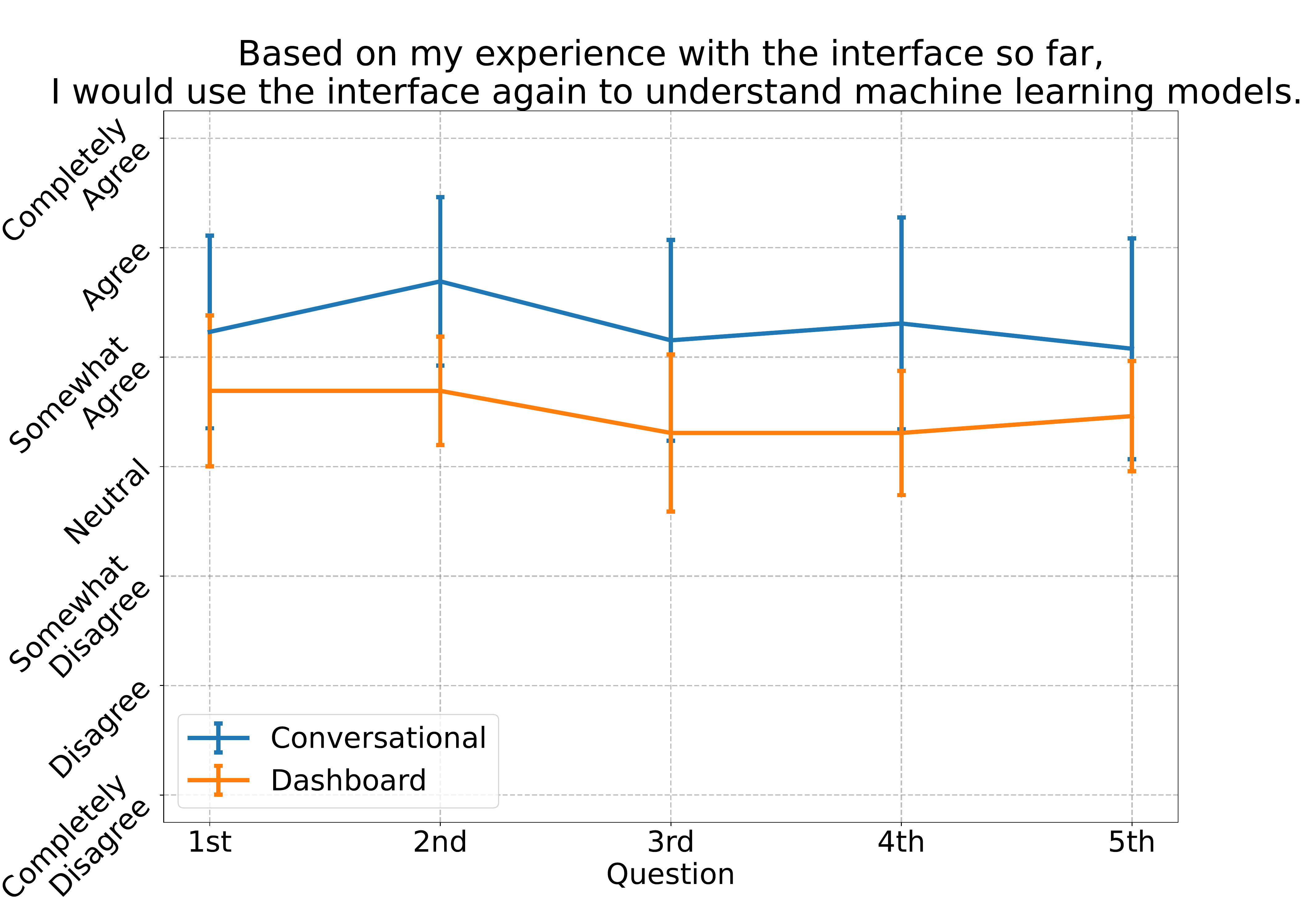}
    \caption{\textbf{ML professionals per question likert results:} these participants were more mixed about which interface they preferred while taking the survey. Interestingly, while the participants were much more accurate using \sys, this population rated their answers at a similar confidence and said they trusted the interfaces similarly while taking the survey. Error bars are 1 standard deviation.}
    \label{fig:ml-grad-per-q}
\end{figure}

In this subsection, we provide additional user study results.
In addition to the questions asked at the end of the survey (Table~\ref{tab:percieved-results} and Table~\ref{tab:accuracy-results}), we also asked users to rate their experiences using both interfaces on a $1$-$7$ Likert while they were taking the survey.
In particular, we asked users how much they agreed with the following statements:
\begin{itemize}
    \item I am confidence I completed my answer correctly.
    \item Completing this task took me a lot of effort.
    \item The interface was useful for completing the task.
    \item Based on my experience so far, I trust the interface to understand machine learning models.
    \item Based on my experience so far, I would use the interface again to understand machine learning models.
\end{itemize}
To evaluate these results, we compute the mean and standard deviation of the Likert score for the $1$st through $5$th question each user sees (question ordering is randomized so users see different questions first).
We compute this for each statement and interface.
The results for the medical workers are provided in Figure~\ref{fig:medical-per-q} and the ML professionals in Figure~\ref{fig:ml-grad-per-q}.
Overall, the medical workers clearly prefer \sys while answering the questions.
Interestingly, they seem to gain trust in \sys over time, going from ``somewhat agree'' to ``agree'' with the statement ``Based on my experience so far, I trust the interface to understand ML models'' by the end of the survey.

\subsection{XAI Question Bank}
\label{subsec:xaiquestionbank}

Here, we provide parses in the \sys for the prototypical questions given in the XAI question bank~\cite{Liao2020QuestioningTA}.
Our grammar can parse $30/31$ core, prototypical questions, excluding socio-technical questions, demonstrating the grammar's broad coverage.
Note, that questions provided in the question bank vary in how they are phrased regarding whether additional coreference is necessary.
For instance, the question bank includes both questions of the form ``what do you predict for this?'' versus ``what do you predict for Q?'').
For conciseness, we write each question in the form where no further coreference is necessary (``what do you predict for Q?'').
For the case where additional coreference is necessary it is straightforward to use the \texttt{previous\_filter} operation to resolve the coreference.
These results demonstrates the \sys grammar is well equipped to support XAI questions.

\begin{table}
    \small
    \centering
    \caption{Prototypical questions from the XAI question bank, parses in the \sys grammar, and explanations of the parse.}
    \label{tab:xai-question-bank1}
    \begin{tabular}{l}
        \toprule
        What kind of output does the system give? \\ 
        \texttt{function()} \\ \vspace{2mm}
        System overview describes model output. \\
        What does the system output mean? \\ \texttt{function()} \\ \vspace{2mm}
        System overview describes meaning of model outputs (e.g., predict if someone has diabetes). \\
        What kind of data was the system trained on? \\ \texttt{data(training\_data)} \\ \vspace{2mm}
        Data overview provides summary of dataset. \\
        What is the sample size of the training data? \\ \texttt{count(training\_data)} \\ \vspace{2mm}
        Count provides number of items in the training data. \\
        What is the distribution of the training data with with a given feature? \\ \texttt{statistic(training\_data, feature\_name)} \\ \vspace{2mm}
        Statistic summarizes feature distribution \\
        How accurate are the predictions? \\
        \texttt{score(test\_data, accuracy)} \\ \vspace{2mm}
        Scoring functionality gets accuracy on test set \\
        How often does the system make mistakes? \\ \texttt{incorrect(test\_data, accuracy)} \\ \vspace{2mm}
        Shows how often the model makes incorrect predictions. \\
        In what situations is the system likely to be incorrect? \\
        \texttt{mistakes(test\_data)} \\ \vspace{2mm}
        Summarizes the common situations mistakes are made and ways the system is wrong. \\
        What kind of mistakes is the system likely to make? \\ \texttt{mistakes(test\_data)} \\ \vspace{2mm}
        Summarizes the common situations mistakes are made and ways the system is wrong. \\
        How does the system make predictions? \\ \texttt{explain(test\_data, feature\_importance}) \\ \vspace{2mm} Provides an overview of features used for making predictions \\
        
        What features does the system consider? \\ \texttt{topk(test\_data, all)} \\ \vspace{2mm} Shows the feature ranking to demonstrate which features the model uses. \\
        
        What would the system predict if a given feature A changes to..? \\
        \texttt{predict(change(filter(test\_data, id, A), feature, value, set))} \\ \vspace{2mm}
        Shows predictions under single feature change. \\
        
        Is feature X used or not used for the predictions \\ \texttt{important(test\_data, X)} \\ \vspace{2mm}
        Determines whether feature X is important for the prediction. \\
        
                How should a given feature A change for this instance to get a different prediction Q? \\
        \texttt{statistic(cfe(filter(test\_data, id, Q, =), 100), summary, A)} \\ \vspace{2mm}
        Summarizes changes to feature that will flip prediction. \\
        
        What is the systems overall logic? \\ \texttt{explain(test\_data, feature\_importance) interaction(test\_data)} \\ \vspace{2mm}
        Provides first order feature importances and second order interaction effects to explain overall system logic. \\
    
        \bottomrule
    \end{tabular}
\end{table}

\begin{table}
    \small
    \centering
    \caption{Prototypical questions from the XAI question bank, parses in the \sys grammar, and explanations of the parse (continued).}
    \label{tab:xai-question-bank2}
    \begin{tabular}{l}
        \toprule
         
        What features of instance Q determine the system's prediction of it? \\ \texttt{topk(filter(test\_data, id, Q, =), all)} \\ \vspace{2mm} Shows most important features that determine prediction. \\
        
        Why are instance A and B given the same prediction? \\ 
        \texttt{explain(or(filter(test\_data, id, A, =), filter(test\_data, id B, =))), feature\_importance)} \\ \vspace{2mm} Explains predictions for both instances by summarizing shared most important features across instances. \\
        
        Why is this instance not predicted to be Q? \\ 
        \texttt{explain(filter(text\_data, id, A, =), feature\_importance, class=Q)} \\ \vspace{2mm}
        Explains alternate class prediction. \\
        
        Why are instance A and B given different predictions? \\
        \texttt{statistic(or(explain(filter(text\_data, id, A, =), feature\_importance),} \\ \quad\quad\texttt{explain(filter(text\_data, id, B, =), feature\_importance)), range, all)} \\ \vspace{2mm}
        Explanations reason for different prediction by contrasting feature importances. \\
        
        How should instance A change to get a different prediction Q? \\
        \texttt{cfe(filter(test\_data, id, A, =), 10, Q)} \\ \vspace{2mm}
        Computes several counterfactual explanations to provide different ways to get alternate predictions. \\
        
        What kind of algorithm is used? \\ \texttt{model()} \\ \vspace{2mm} Describes the ML model. \\
        Why is the instance given this prediction? \\ \texttt{explain(filter(test\_data, id, X, =), feature, feature\_importance)} \\ \vspace{2mm} Explains prediction for given instance with feature importance.\\
        
        What is the minimum change required for instance A to get a different prediction Q? \\
        \texttt{cfe(filter(test\_data, id, A, =), 1, Q)} \\ \vspace{2mm}
        Computes single, minimal counterfactual. \\
        
        What kind of instance is predicted outcome Q? \\
        \texttt{statistic(filter(test\_data, Y, Q, =), summary, all)} \\ \vspace{2mm}
        Summarizes instances with certain predicted outcome. \\
        
        What is the scope of change permitted for instance A to still get the same prediction? \\
        \texttt{statistic(cfe(filter(test\_data, id, A, =), 100), min, all)} \\ \vspace{2mm}
        Summarizes minimal changes to flip prediction. \\
        
        What is the range of values permitted to for a given feature for this prediction  on A to stay the same? \\
        \texttt{statistic(cfe(filter(test\_data, id, A, =), 100), min, feature)} \\ \vspace{2mm}
        Summarizes minimal changes to certain feature that will flip prediction. \\
         
        What kind of instance gets the same prediction Q? \\
        \texttt{statistic(filter(test\_data, Y, Q, =))} \\ \vspace{2mm}
        Summarizes instances with the same prediction Q. \\
        
        What would the system predict if instance A changes to...? \\
        \texttt{predict(and(change(filter(test\_data, id, A), feature, value, set),}\\ \quad\quad\texttt{change(filter(test\_data, id, A), feature, value, set)...))} \\ \vspace{2mm}
        Shows predictions under potentially many different changes to instance. \\
        
         What would the system predict for [a different instance A]? \\
         \texttt{predict(filter(test\_data, feature, A, =))} \\ \vspace{1mm}
         Shows predictions for different instances according to filtering criteria. \\
         
         What does [a machine learning terminology] mean? \\
         \texttt{define(term)} \\  \vspace{1mm}
         Defines a term. \\
        \bottomrule
    \end{tabular}
\end{table}

\clearpage

\section{Interface Screenshots} In this appendix, we provide additional interface screen shots for both \sys and the dashboard baseline.

\subsection{\sys Interface} 
\begin{figure}[h!]
    \centering
    \includegraphics[width=0.7\columnwidth]{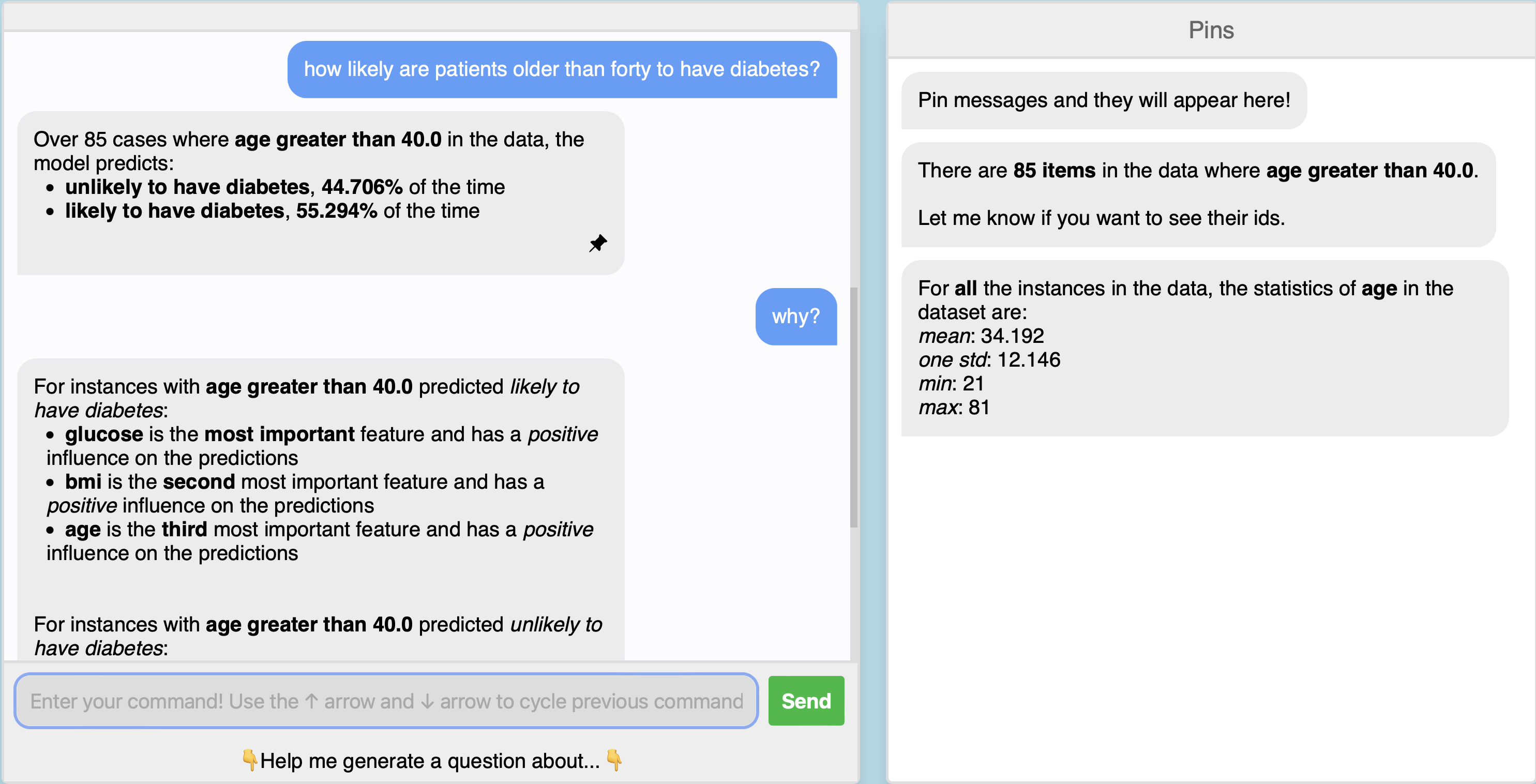}
    \caption{Discussing the model's predictions on patients older than $40$. The system correctly understand's the context of the question ``why?''.}
    \label{fig:ttm1}
\end{figure}

\begin{figure}[h!]
    \centering
    \includegraphics[width=0.7\columnwidth]{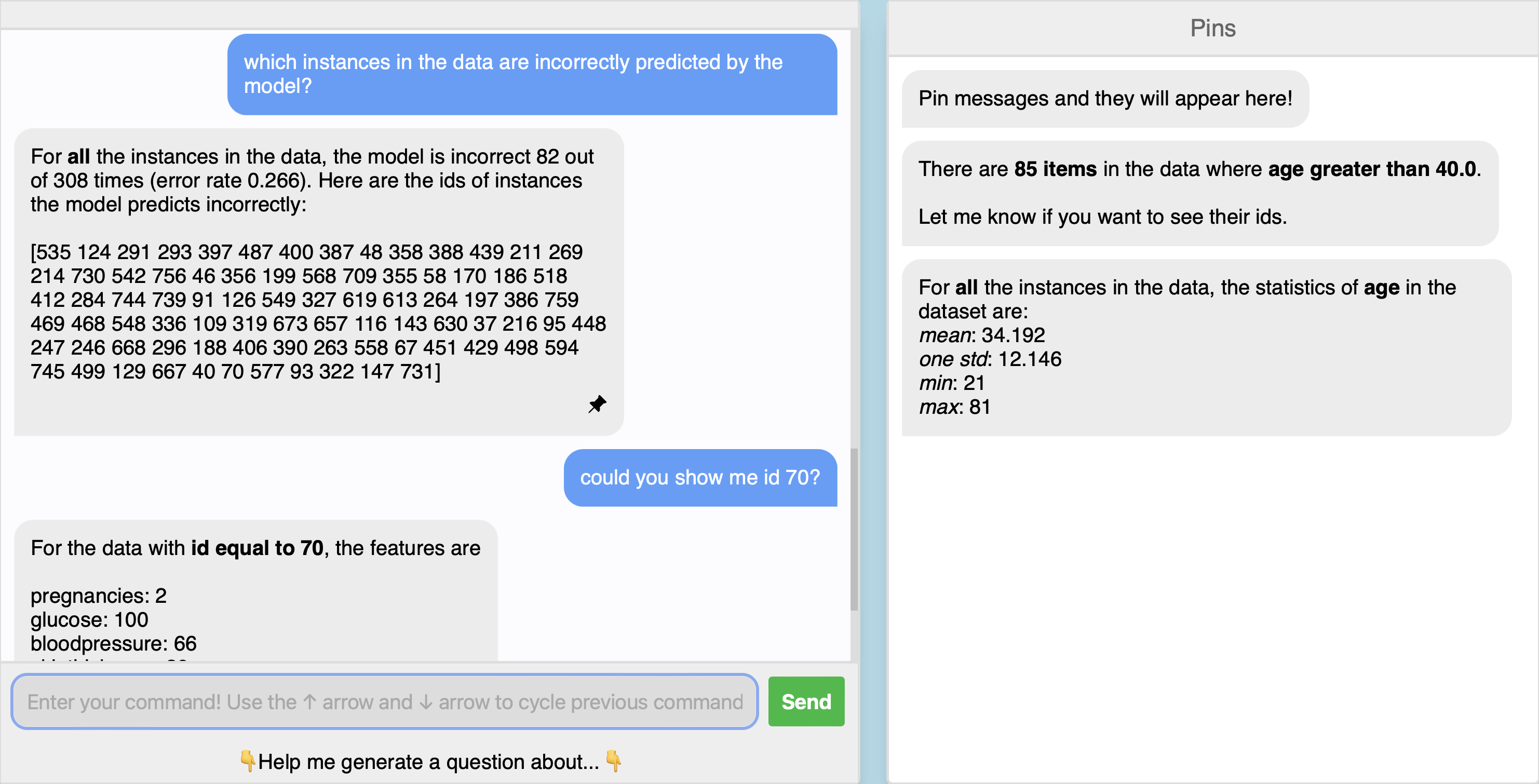}
    \caption{Beyond explanations, \sys is also highly useful for classic error analysis, such as inspecting incorrectly predicted instances, as is shown here.}
    \label{fig:ttm2}
\end{figure}

\begin{figure}[h!]
    \centering
    \includegraphics[width=0.7\columnwidth]{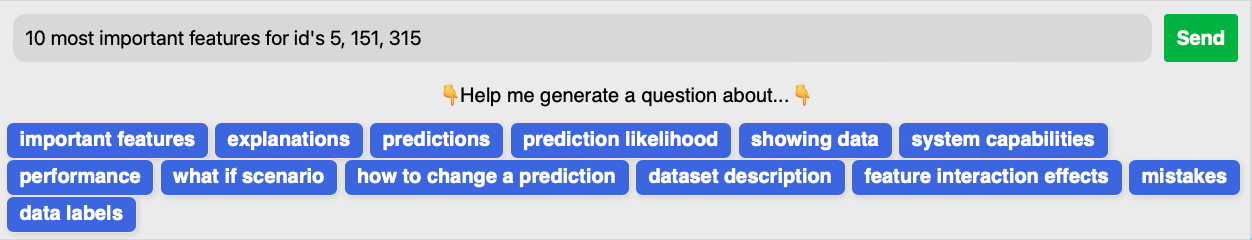}
    \caption{Screen shot of the help generate a question button, which will input a random question from the training data in the category when clicked..}
    \label{fig:genscreenshot}
\end{figure}

We provide additional screenshots of the \sys GUI interface in Fig.~\ref{fig:ttm1} and Fig.~\ref{fig:ttm2}.
The interface is a text chat interface, where users provide input and the system response.
On the right side of the interface, users can ``pin'' messages they find interesting. 
In addition, we provide a feature where users can get help to generate a question in a particular category (Fig.~\ref{fig:genscreenshot}).
This feature inputs a random synthetic training utterance into the command bar from the category of choice.

\clearpage

\subsection{Dashboard Interface} 
\begin{figure}[h!]
    \centering
    \includegraphics[width=0.7\columnwidth]{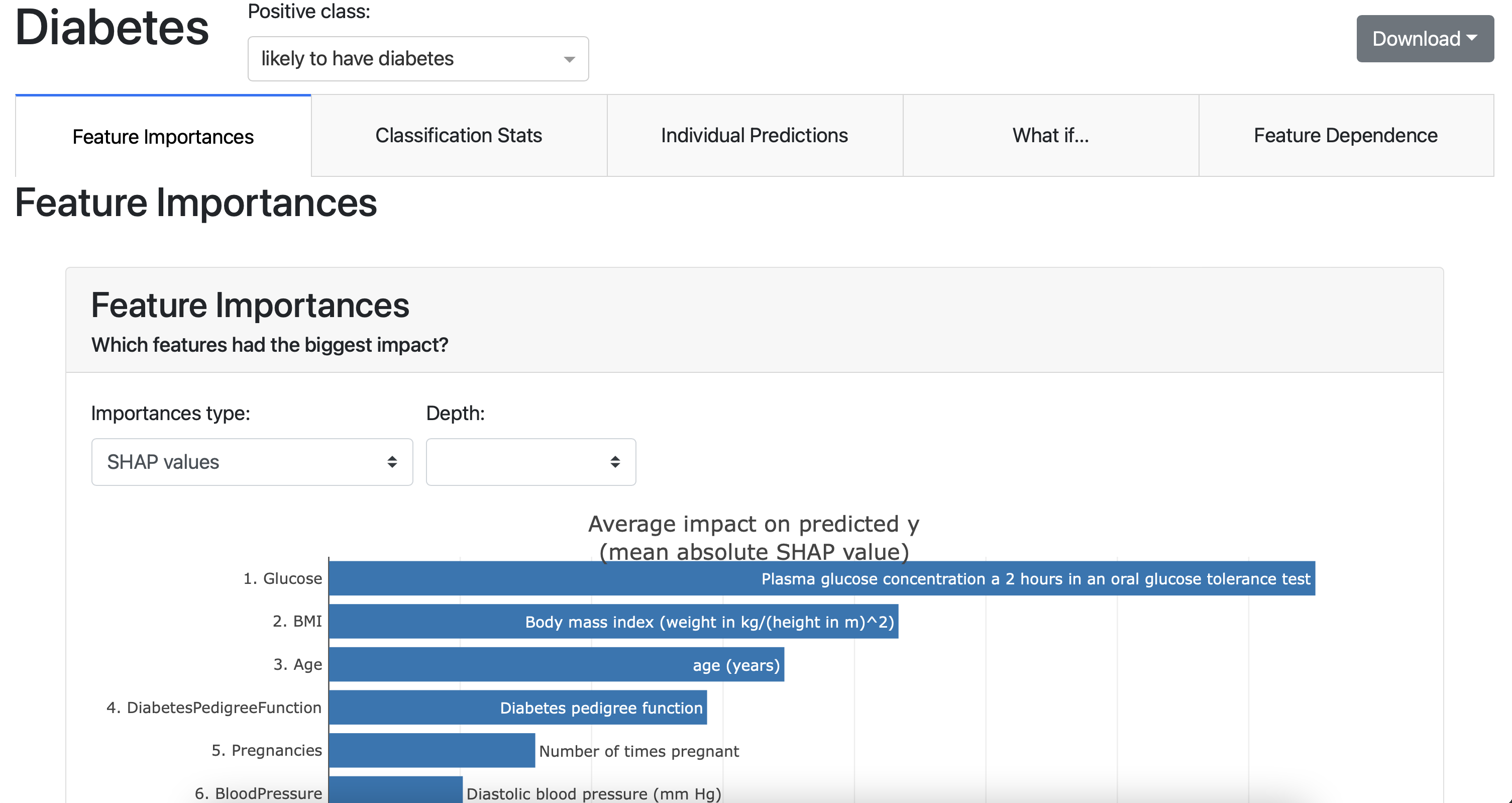}
    \caption{The dashboard interface lands on the global feature importances. Users can navigate to other tabs in the interface to see model predictions, metrics, feature importances, and compute what-if scenarios.}
    \label{fig:dashboard-interface}
\end{figure}

\begin{figure}[h!]
    \centering
    \includegraphics[width=0.7\columnwidth]{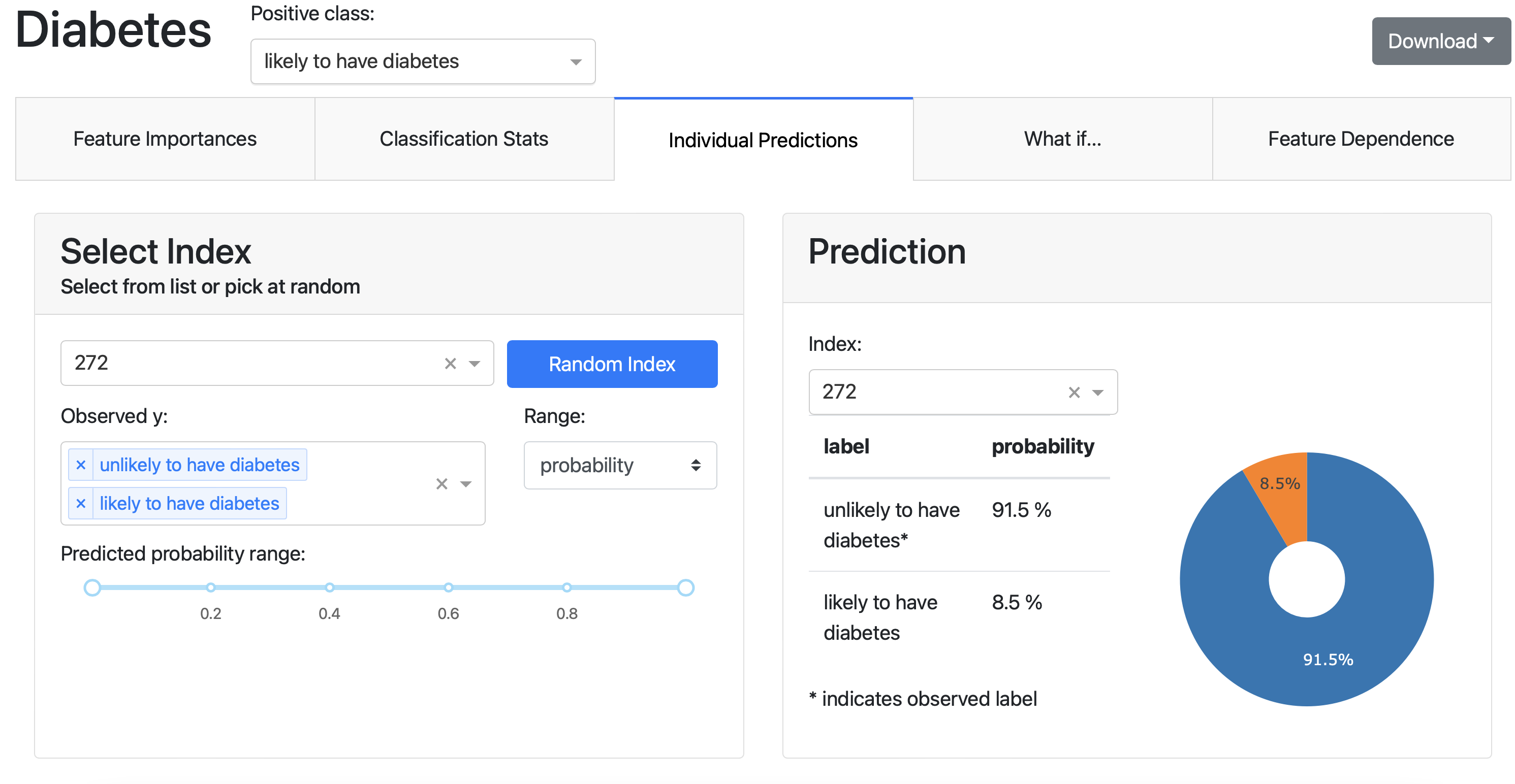}
    \caption{An example of the ``individual predictions'' page in the dashboard interface. Users need to navigate to this page and type in the data point index they want in order to inspect predictions.}
    \label{fig:dashboard-interface2}
\end{figure}

We provide example screenshots for the baseline dashboard interface in Fig.~\ref{fig:dashboard-interface} and Fig.~\ref{fig:dashboard-interface2} ~\cite{oege_dijk_2022_6408776}.
In depth description can be found in the project's documentation \href{https://explainerdashboard.readthedocs.io/en/latest/}{https://explainerdashboard.readthedocs.io/en/latest/}.
The dashboard provides different ways to understand ML models using point and click interactions.

\clearpage

\end{document}